\documentclass[11pt]{article}

\usepackage[a4paper,margin=1in]{geometry}
\usepackage[utf8]{inputenc}
\usepackage[T1]{fontenc}
\usepackage{lmodern}
\usepackage{microtype}
\usepackage{amsmath,amssymb,amsfonts,amsthm,mathtools,bm}
\usepackage{graphicx,booktabs,placeins}
\usepackage[font=small]{caption}
\usepackage{needspace}
\usepackage{xcolor}
\usepackage[round,authoryear]{natbib}

\usepackage{xurl}
\usepackage{hyperref}
\ifdefined\pdfminorversion\pdfminorversion=7\fi
\hypersetup{
  colorlinks=true,
  linkcolor={red!70!black},
  citecolor={blue!70!black},
  urlcolor={red!70!black},
  pdfauthor={Fabrizio Boncoraglio and Lenka Zdeborov\'a},
  pdftitle={When Is Early Stopping Beneficial Under Label Noise? A Phase Diagram for Kernel Gradient Flow}
}
\graphicspath{{Figures/}}

\newcommand{\R}{\mathbb R}
\newcommand{\E}{\mathbb E}
\newcommand{\Prob}{\mathbb P}
\newcommand{\Id}{\mathrm I}
\newcommand{\N}{\mathcal N}
\newcommand{\ind}{\mathbf 1}
\newcommand{\Egen}{\mathcal E_{\rm gen}}
\newcommand{\Emem}{\mathcal E_{\rm mem}}

\newcommand{\bc}{b_{\rm c}}
\newcommand{\tes}{t_{\rm es}}
\newcommand{\tmem}{t_{\rm mem}}
\newcommand{\taue}{\tau_{\rm es}}
\newcommand{\taum}{\tau_{\rm mem}}
\newcommand{\Tr}{\operatorname{Tr}}
\newcommand{\sign}{\operatorname{sign}}
\newcommand{\erfc}{\operatorname{erfc}}

\newcommand{\Var}{\operatorname{Var}}

\newcommand{\op}{\mathrm{op}}

\title{When is early stopping beneficial under label noise? \\ A phase diagram for kernel gradient flow}

\author{
Fabrizio Boncoraglio$^1$,
Lenka Zdeborov\'a$^1$
}

\date{
\small
$^1$ Statistical Physics of Computation Laboratory,
\\
École Polytechnique Fédérale de Lausanne (EPFL)
CH-1015 Lausanne
}

\begin{document}

\maketitle

\begin{abstract} Early stopping is commonly used to prevent fitting label noise from degrading generalization. Does this mean that, when a positive fraction of the training labels is intrinsically unpredictable, the best test error must occur before interpolation? Experiments on a CIFAR10 classification task with randomly labeled examples motivate the opposite possibility: although fixed-feature and shallow models exhibit late-time deterioration of the test error, the need for early stopping essentially disappears in larger, deeper convolutional networks even after the random labels have all been classified correctly. We determine when late-time deterioration disappears entirely in a solvable setting. For kernel gradient flow on a synthetic noisy classification task, we derive sharp high-dimensional learning curves and obtain a phase diagram controlled by the balance between the linear kernel component that learns the shared relation between inputs and labels and the higher-order components that enable fitting the label noise. In one regime, generalization is optimized at finite time and early stopping is necessary. In another, generalization improves until interpolation, but the endpoint is suboptimal across kernels. At the boundary, the interpolating endpoint attains the lowest generalization error across the considered kernel family and all training times. Thus, even with extensive irreducible label noise, interpolation can outperform early stopping. The neural-network experiments leave open how learned representations acquire an analogous favorable geometry. \end{abstract}

\begin{figure}[!htbp] \centering \includegraphics[width=\linewidth]{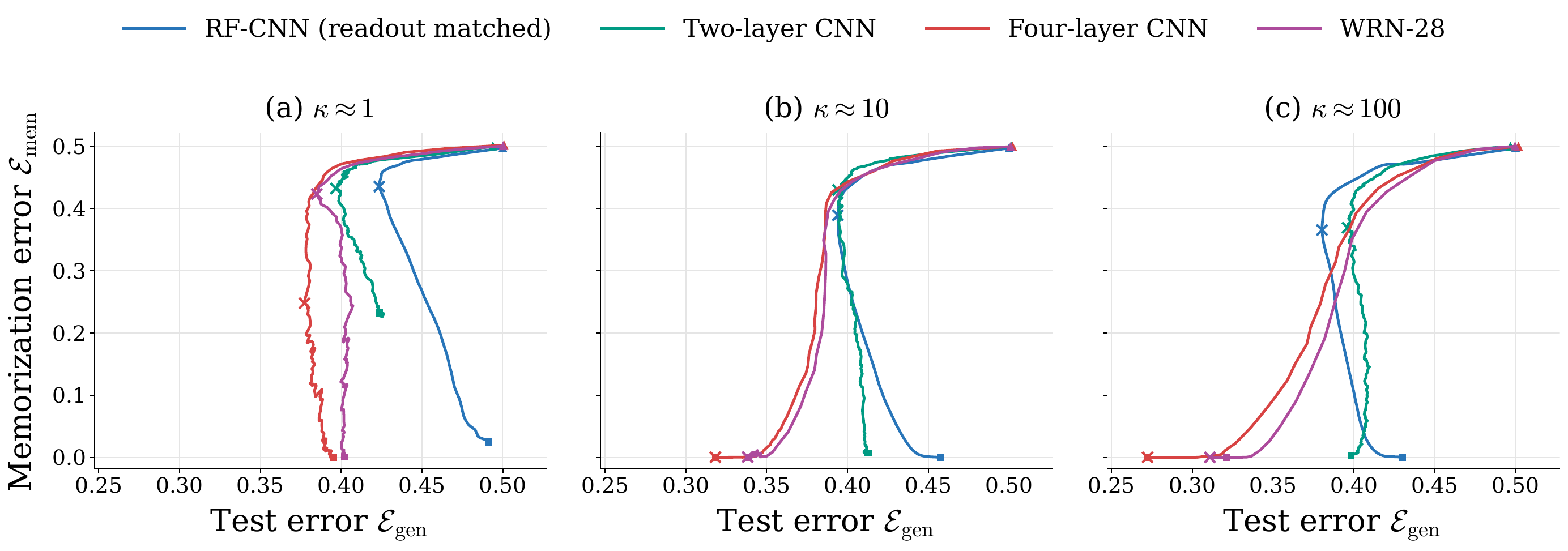}

\caption{\textbf{Generalization along the path to memorization.} Test error on 2000 cats and dogs ($x$-axes) versus classification error on the randomly labelled automobiles ($y$-axes) taken from the CIFAR10 dataset. The fixed training set has $n=7000$ images: $2800$ cats, $2800$ dogs with their respective labels, and $1400$ automobiles with independent random labels. Panels vary the overparametrization ratio $\kappa=P/n\approx1,10,100$, where $P$ counts optimized parameters. Curves average about 30 completed seeds; a centered arithmetic average of up to seven neighboring recorded times is applied to both coordinates for display, preserving the initial and final values. These parametric curves trace the two errors as training time $t$ increases. Triangles mark initialization, squares the last iterate and crosses the minimum displayed test error. All trainable networks use square loss and constant-rate Nesterov SGD, without augmentation or weight decay, through 1500 epochs; RF-CNN uses full-batch readout GD through $10^{12}$ steps. In Appendix~\ref{app:nn} we provide all numerical details and the unsmoothed trajectories, including training loss.}

\label{fig:capacity} \end{figure}

\section{Introduction}

\label{sec:introduction}

Early stopping is a standard mechanism used to prevent overfitting, in particular in the presence of label noise. The usual dynamical picture is that training first learns structure shared across examples and only later fits labels that cannot be predicted from the inputs. Continued training then degrades test performance, so the best generalization error is attained at a finite time, before interpolation. This intuition is made precise in theoretical analyses of linear, kernel, and neural-network dynamics, where early stopping acts as an implicit regularizer that prevents the learner from fitting corrupted labels \citep{yao2007,raskutti2014,ali2019,li2020}. One might therefore expect early stopping to be particularly important when an extensive fraction of the training labels is intrinsically unpredictable.

We challenge this expectation through a controlled experiment based on CIFAR10 \citep{krizhevsky2009}. Cats and dogs define a shared classification task, while automobile images, comprising 20\% of the training set, receive fixed random labels. We measure the generalization error on new samples of cats and dogs and the memorization error, defined as the fraction of misclassified automobiles in the training set, to track when the unpredictable labels have been fitted. Figure~\ref{fig:capacity} follows the two errors jointly as training progresses across architectures and model sizes. Fixed-feature and smaller or shallower models exhibit the familiar behavior: as the memorization error decreases, the test performance eventually deteriorates. For the larger, deeper convolutional and residual networks, this deterioration is essentially absent: their best observed test performance is attained with zero memorization error, and early stopping does not visibly improve generalization.

\begin{center} \fbox{\parbox{0.92\columnwidth}{ \textbf{Empirical hypothesis.} In image classification, even with extensive label noise, large, deep neural networks with competitive test performance do not benefit from early stopping. }} \end{center}

Appendix~\ref{app:nn} examines this empirical hypothesis in more detail. We report the unsmoothed neural network trajectories and training losses behind Figure~\ref{fig:capacity}, together with additional convolutional and residual architectures. We also consider other training procedures commonly used to obtain stronger performance on image-classification tasks, combining cross-entropy loss, data augmentation, learning-rate warm-up and cosine decay, followed by an extended constant-rate tail. We also replace the automobiles by randomly labelled cats and dogs, so that clean and randomized examples belong to the same image classes. The evidence collected in Appendix~\ref{app:nn} robustly points to the same qualitative behavior: the larger and more complex networks attain their best observed test performance after all random labels have been classified correctly, with little subsequent deterioration. Further controls examine the effects of the optimizer, batch size, training-set size and fraction of random labels.

This raises the theoretical question of whether there are settings in which one can show analytically that training all the way to interpolation gives the lowest test error.

This question is related to, but distinct from, grokking and benign overfitting. Canonical grokking concerns noiseless, realizable tasks, where training accuracy becomes essentially perfect long before test performance improves \citep{power2022}. Here, instead, a considerable fraction of the labels is intrinsically unpredictable, and we ask whether fitting them eventually harms generalization. Benign overfitting asks whether an interpolating estimator can generalize well despite label noise \citep{bartlett2020benign,liang2020just}; it does not by itself determine whether interpolation is optimal relative to earlier points along the training trajectory. Our focus is precisely on this dynamical comparison between early stopping and convergence.

Can the minimum test error nevertheless genuinely occur at convergence? Existing theoretical analyses that follow feature-learning dynamics under label noise toward interpolation reach the opposite conclusion for the shallow neural networks they study in the regimes analyzed: fitting the noisy labels eventually degrades generalization, making early stopping preferable \citep{han2025role,montanari2026dynamical}. They therefore do not account for the empirical hypothesis formulated here. We show that such behavior is possible for square-loss kernel gradient flow in the Rules-and-Facts (RAF) model introduced by \citet{farne2026}.

Each training label either follows a shared linear rule or is independent of the input. Whereas \citet{farne2026} analyze the resulting empirical risk minimizers (ERM) under explicit regularization, here we follow the full learning trajectory from initialization to interpolation in the absence of explicit regularization. In the proportional high-dimensional limit, this trajectory is governed by the relative strengths of the rule-aligned component and the higher-order kernel components that provide the capacity to fit the unpredictable labels.

\paragraph{Main theoretical result.} We derive exact learning curves and obtain a phase diagram with three regimes, corresponding to three families of kernels. In the first, test error reaches its minimum at finite time and subsequently increases, so early stopping is necessary. In the second, test error improves until convergence, but the resulting interpolating predictor is suboptimal relative to other kernel geometries. Precisely at the boundary between these regimes, gradient flow converges to an interpolating predictor that attains the lowest generalization error across the considered kernel families and all training times. Thus, even in the presence of extensive label noise, interpolation can lead to a better test error than any early stopping.

The kernel analysis therefore produces the behavior that motivated this work. At the critical geometry, the best test error is attained at interpolation, and the corresponding trajectory in Figure~\ref{fig:theory-parametric} closely resembles those of the larger, deeper neural networks in Figure~\ref{fig:capacity}, whose best test performance is attained after all randomly labelled examples have been classified correctly. The theory thus provides a concrete realization of the observed phenomenon. It does not, however, explain how neural-network training constructs a representation with analogous properties. Determining which aspects of feature learning, model size, and architecture produce this behavior remains an open question.

\subsection*{Related work.}

\paragraph{Learning with label noise.} Deep networks can fit arbitrary random labels \citep{zhang2017}, while often learning shared structure before memorizing individual samples \citep{arpit2017}. This separation of time scales motivates early stopping as a method for improving robustness to label noise \citep{li2020,bai2021}. The severity of late-time overfitting, however, depends on model capacity and optimization, as illustrated by epoch-wise double descent \citep{nakkiran2020,heckel2021} and by the weak overtraining observed in sufficiently overparameterized models \citep{advani2020}. These works do not provide an exact criterion for when fitting the remaining label noise ceases to harm generalization.

\paragraph{Early stopping and learning trajectories.} For least squares, early stopping acts as a spectral regularizer \citep{yao2007,raskutti2014}, and the complete gradient-flow trajectory has an explicit matrix-exponential form \citep{ali2019}. High-dimensional learning curves have been derived for linear, kernel, and Gaussian-covariate models \citep{advani2020,ghosh2022,bodin2022}, including regimes in which training error decreases while test error remains essentially unchanged \citep{ghosh2022}. Recent work also interprets early stopping through a transient spectral transition in an anisotropic linear model \citep{coeurdoux2026}.

These analyses do not characterize the transition between lowest test error at finite time and at convergence across kernels.

\paragraph{Early stopping and explicit regularization.} Early stopping, Tikhonov regularization and other spectral filters can attain comparable statistical rates \citep{bauer2007,yao2007,raskutti2014,celisse2021analyzing,hu2022}. For least-squares regression, \citet{ali2019} compare gradient flow with ridge regression under the calibration $\lambda\asymp1/t$ and obtain uniform constant-factor risk comparisons. More recent work shows that a finitely stopped gradient-descent iterate can be represented as a generalized, spectrally dependent ridge solution \citep{sonthalia2024}, while high-dimensional and finite-sample comparisons demonstrate that optimally stopped gradient descent and optimally tuned scalar ridge need not attain the same risk \citep{stark2025,wu2025risk}. Our setting makes this distinction fully explicit for classification error.

\paragraph{Benign overfitting and kernel geometry.} Benign-overfitting theory characterizes when an interpolating estimator can generalize well despite label noise \citep{bartlett2020benign,liang2020just}. More generally, kernel spectra and their alignment with the target control generalization \citep{bordelon2020,canatar2021,bordelon2022}, while high-dimensional kernel linearization makes this dependence explicit for broad kernel families \citep{elkaroui2010,dovu2013}. The RAF model provides a characterization of how kernel geometry allocates capacity between rule learning and fitting unpredictable labels \citep{farne2026}. These results do not compare full learning trajectories across kernel geometries to determine when the test error at interpolation is lower than at every finite stopping time.

\section{Data \& kernel dynamics}

\label{sec:setting} \paragraph{Synthetic data model.} The Rules-and-Facts (RAF) model proposed in \citet{farne2026} separates a shared prediction rule from individual associations, which we treat as simple label noise for this paper. We draw independent inputs uniformly on the sphere of radius $\sqrt d$, namely $x^\mu=\sqrt d\,z^\mu/\|z^\mu\|_2$ with $z^\mu\sim\N(0,\Id_d)$. An independent ground-truth vector $w_\star\sim\N(0,\Id_d)$ defines $g_\star(x)=w_\star^\top x/\sqrt d$. Training labels are given by 
\begin{equation} 
\begin{gathered} 
y^\mu=\begin{cases}\sign g_\star(x^\mu),&c^\mu=1,\\ \eta^\mu,&c^\mu=0,\end{cases}\\ c^\mu\sim\operatorname{Bernoulli}(1-\varepsilon),\qquad \Prob(\eta^\mu=+1)=\Prob(\eta^\mu=-1)=\tfrac12. \end{gathered} 
\label{eq:data} 
\end{equation}

Here $0<\varepsilon<1$, so a fraction $\varepsilon$ of the training labels is independent of the inputs, while the remaining fraction $1-\varepsilon$ follows the shared rule $w^\star$. The assignments and random labels are drawn once and remain fixed throughout training. The learner observes only the supervised dataset $\mathcal D=\{(x^\mu,y^\mu)\}_{\mu=1}^n$. This construction provides a controlled setting in which we can follow the fitting of intrinsically unpredictable labels separately from generalization of the shared rule.

For a given predictor $f_t(x)$, we measure rule learning and factual memorization through the errors \begin{equation} 
\begin{aligned} 
\Egen(t)&=\E_x\ind\{\sign f_t(x)\ne\sign g_\star(x)\},
\label{eq:gen-def} 
\end{aligned} 
\end{equation}

\begin{equation} 
\begin{aligned} 
\Emem(t)&=\frac1{|\mathcal I_{\rm mem}|}\sum_{\mu\in\mathcal I_{\rm mem}} \ind\{\sign f_t(x^\mu)\ne y^\mu\},\\ &\quad \mathcal I_{\rm mem}=\{\mu:c^\mu=0\}.
\label{eq:mem-def} 
\end{aligned} 
\end{equation}

Generalization concerns a fresh input with its rule label; the quantity $\mathcal E_{\rm mem}$ is primarily a diagnostic of when training has fitted the unpredictable labels. In the kernel analysis, interpolation is achieved for zero empirical square loss; this is stronger than $\Emem=0$. For the unregularized dynamics below, the positive-definite kernel reaches interpolation as $t\to \infty$; we ask whether the lowest test error occurs at a finite time or at this endpoint. Appendix~\ref{app:nn} reports classification errors and losses separately for the neural-network experiments.

\paragraph{Kernel gradient flow.} We consider the kernel predictor $f_t(x)=\sum_\mu a_{t\mu}k(x^\mu,x)$, and write $K_{\mu\nu}=k(x^\mu,x^\nu)$ and $k(x)_\mu=k(x^\mu,x)$. Here $k(x,x')$ is a fixed positive-semidefinite similarity kernel on inputs in $\R^d$, $K=(K_{\mu\nu})_{\mu,\nu=1}^n$ is its training matrix, and $k(x)\in\R^n$ compares the training inputs with a new input $x$. We use standard kernel gradient flow \citep{ghosh2022}: for square loss this gives 
\begin{equation} 
\begin{aligned} 
L_\lambda(\bm a)&=\tfrac12\|y-K\bm a\|_2^2 +\tfrac\lambda2\bm a^\top K\bm a,\\ \dot{\bm a}_t&=y-(K+\lambda\Id_n)\bm a_t. 
\end{aligned} 
\label{eq:flow} 
\end{equation} 
The coefficient vector is $\bm a_t=(a_{t1},\ldots,a_{tn})^\top\in\R^n$, and $y=(y^1,\ldots,y^n)^\top$ is the vector of training labels. The kernel is fixed: the gradient is taken with respect to the function $f$ in its reproducing-kernel Hilbert space (RKHS), rather than with respect to the coefficient vector $\bm a$. Both lead to analogous results.

Appendix~\ref{app:metric} discusses the alternative Euclidean flow dynamics and shows that it preserves the early-stopping boundary we derive below. Kernel gradient flow also describes the fixed-kernel dynamics associated with linearized wide neural networks \citep{jacot2018} .

Given an initialization $\bm a_0\sim\N(0,\Delta_0\Id_n)$ independent of the data and labels, Eq.~\eqref{eq:flow} admits the standard matrix-exponential solution \begin{equation} 
\begin{aligned} 
\bm a_t&=R_t\bm a_0 \quad +(K+\lambda\Id_n)^{-1}(\Id_n-R_t)y,\\ R_t&=e^{-t(K+\lambda\Id_n)},\quad f_t(x)=k(x)^\top\bm a_t. 
\end{aligned} 
\label{eq:finite-flow} 
\end{equation}

When \(K\succ0\) and \(\lambda=0\), the training predictions satisfy \(K\bm a_t\to y\), and hence \(L_0(\bm a_t)\to0\) as \(t\to\infty\) . The limiting predictor is therefore the interpolating endpoint of the trajectory. Our main results concern this unregularized dynamics; we retain fixed \(\lambda\geq0\) in intermediate formulas to state the corresponding extension.

The central dynamical quantity is the time of best generalization, 
\begin{equation} 
t_{\rm es} \in \operatorname*{arg\,min}_{t\in[0,\infty]} \mathcal{E}_{\rm gen}(t). 
\label{eq:optimal_stopping_time} 
\end{equation} 
The subscript ``es'' stands for early stopping. We write \(t_{\rm es}=\infty\) when the minimum is attained only at convergence. This is an oracle quantity describing the trajectory; selecting a stopping time in practice would require validation data.

\paragraph{High-dimensional kernel reduction.} We consider the high-dimensional limit $n,d\to\infty$ with $n/d\to\alpha\in(0,\infty)$. In this work we consider positive-semidefinite dot-product kernels $k(x,x')=\sigma(x^\top x'/d)$ with $\sigma(\cdot)$ fixed and with three continuous derivatives near zero and a finite value at one. Define 
\begin{equation} 
\begin{aligned} 
\mu_0^2&=\sigma(0)\ge0,\\ \mu_1^2&=\sigma'(0)>0,\\ \mu_\star^2&=\sigma(1)-\sigma(0)-\sigma'(0)>0. \end{aligned} 
\label{eq:coefficients} 
\end{equation} 
With $X$ having rows $x^{\mu\top}$ and $C_n=XX^\top/d$, kernel linearization \citep{elkaroui2010,dovu2013} yields \begin{equation} 
\begin{aligned} 
K=\left[\mu_0^2+\frac{\sigma''(0)}{2d}\right]\bm1\bm1^\top +\mu_1^2C_n+\mu_\star^2\Id_n+R_d. 
\end{aligned} 
\label{eq:kernel} 
\end{equation} 
For every $u>0$, $\Prob(\|R_d\|_{\op}>u)\to0$ as $n,d\to\infty$ with $n/d\to\alpha$; the kernel and initialization variance remain fixed. Although the rank-one term matters for this matrix identity, we show in Appendix~\ref{app:kernel} that it does not affect the limiting memorization and generalization errors at fixed positive time. We can therefore compute the errors using the equivalent kernel $K_{\rm eff}=\mu_1^2C_n+\mu_\star^2\Id_n$. The assumption $\mu_\star^2>0$ excludes purely linear kernels. We will show that only the ratio of $\mu_1$ and $\mu_\star$ enters the error trajectories, after appropriate rescalings.

This reduction exposes the kernel geometry relevant to our question. The component $\mu_1^2C_n$ retains the input-dependent directions that support learning the shared rule. The diagonal component $\mu_\star^2I_n$, which aggregates the higher-order part of the nonlinear kernel, namely $\sigma(1)-\sigma(0)-\sigma'(0)$, provides the additional capacity needed to fit unpredictable training labels. Their relative strength determines whether continued training improves or eventually harms generalization, while the overall scale only sets the training-time scale. In all numerical validations, we use the original nonlinear kernel rather than imposing the linearized form.

\section{Learning curves and the early-stopping boundary}

\label{sec:errors} We now determine when kernel gradient flow should be stopped before convergence. In the proportional high-dimensional limit, we derive the complete generalization and memorization trajectories. Although the shape and time scale of these trajectories depend on the sample ratio, initialization, and overall kernel scale, whether generalization is optimized at finite time or at convergence is governed by a single kernel-geometry parameter. We first state the exact learning curves and then identify this early-stopping boundary.

At each fixed positive reduced time $\tau=\mu_1^2t$ and in the high-dimensional limit stated above, the two errors in Eqs.~\eqref{eq:gen-def} and \eqref{eq:mem-def} converge to \begin{equation} 
\begin{aligned} \Egen(\tau)&=\frac1\pi\arccos\!\left(\frac{m_\tau}{\sqrt{q_\tau}}\right),\\ \Emem(\tau)&=\frac12\erfc\!\left(\frac{V_\tau}{\sqrt{2Q_\tau}}\right). 
\end{aligned} 
\label{eq:errors} 
\end{equation} 
The four quantities $m_\tau,q_\tau,V_\tau,Q_\tau$ are time-dependent scalars. The finite optimal stopping time defined in Eq.~\eqref{eq:optimal_stopping_time} is found by solving the one-dimensional stationarity condition minimizing $t_{\rm es}\in (\arg\min_\tau \Egen(\tau))/\mu_1^2$, with $\taue=\mu_1^2\tes$, and comparing the generalization errors at its candidate minima and at the endpoints.

The scalar parameters $m_\tau,q_\tau,V_\tau,Q_\tau$ satisfy the closed one-dimensional expressions 
\begin{equation} 
\begin{aligned} 
m_\tau&=a\alpha\langle F_\tau\rangle_\rho,\\ q_\tau&=\alpha\left\langle \left(a^2+\frac{1-a^2}{s}\right)F_\tau^2+\delta_0sr_\tau^2 \right\rangle_\rho,
\label{eq:mq} 
\end{aligned} 
\end{equation}

\begin{equation} 
\begin{aligned} 
V_\tau&=\langle h_\tau\rangle_\rho,\\ Q_\tau&=\left\langle(1-a^2+a^2s)(h_\tau-V_\tau)^2 +\delta_0(s+b)^2r_\tau^2\right\rangle_\rho, 
\label{eq:VQ} 
\end{aligned} 
\end{equation} 
where 
\begin{equation} 
\begin{aligned} 
r_\tau(s)&=e^{-\tau(s+b+\ell)},\\ h_\tau(s)&=\frac{s+b}{s+b+\ell}(1-r_\tau(s)),\\ F_\tau(s)&=\frac{s}{s+b+\ell}(1-r_\tau(s)). 
\end{aligned} 
\label{eq:filters} 
\end{equation} 

Here the variable $s$ represents an eigenvalue of $C_n=XX^\top/d$; the corresponding effective-kernel eigenvalue is $\mu_1^2s+\mu_\star^2$. We write $\langle\psi\rangle_\rho=\int\psi(s)\rho(ds)$ for an average over the limiting eigenvalues of $C_n$. For spherical inputs, $\rho$ is the Marchenko--Pastur law and $s$ the corresponding random variable \citep{marchenko1967}, with details in Appendix~\ref{app:derivations}; $F_\tau^2/s$ has continuous value zero at $s=0$. The remaining variables in Eqs. \eqref{eq:mq} and \eqref{eq:VQ} are 

\begin{equation} 
\begin{aligned} 
a&=(1-\varepsilon)\sqrt{2/\pi}, \quad \gamma=\arctan\frac{\mu_1}{\mu_\star} ,\\ \ell&=\frac\lambda{\mu_1^2},\quad \delta_0=\mu_1^4\Delta_0,\quad b=\frac{\mu_\star^2}{\mu_1^2}=\cot^2\gamma. 
\end{aligned} 
\label{eq:reduced} 
\end{equation} 

Thus, at fixed reduced ridge $\ell$ and initialization $\delta_0$, kernel geometry enters only through the kernel angle~$\gamma$. Different kernels are therefore distinguished by $\gamma$, and kernels with the same angle have identical error trajectories in reduced time. In the formulas above $m_\tau=\lim_{d\to\infty}\E_x[f_t(x)g_\star(x)]$ and $q_\tau=\lim_{d\to\infty}\E_x[f_t(x)^2]$ measure alignment with the rule and variance of predictions on new inputs.

The quantities $V_\tau,Q_\tau$ are instead the limiting mean and variance of the signed margin $y^\mu f_t(x^\mu)$ for a typical training fact. Its asymptotic Gaussian law gives the memorization error in Eq.~\eqref{eq:errors}. All time limits below are taken after the high-dimensional limit.

The learning curves are obtained by solving the gradient-flow dynamics independently in each eigenvector of $C_n$, as is well known from \citep{ali2019,advani2020,bodin2022}; all the details are given in Appendix~\ref{app:derivations}. The functions $r_\tau$, $h_\tau$ and $F_\tau$ in Eq.~\eqref{eq:filters} are usually called spectral filters: they determine how each eigenvector of $C_n$ contributes to the prediction at time $\tau$ \citep{ali2019}. To compute $\Emem(\tau)$, we additionally derive the asymptotic distribution of the signed training margins on the randomly labelled examples, whose mean and variance are $V_\tau$ and $Q_\tau$. At $\lambda=0$, gradient flow converges to exact interpolation, meaning zero squared loss on the training set. Correspondingly, $h_\infty=1$, and hence $V_\infty=1$, $Q_\infty=0$, and $\Emem(\infty)=0$. Interpolation alone, however, does not determine generalization: the spectral filter entering test predictions is $F_\infty(s)=s/(s+b)$ and therefore depends on the kernel geometry~$\gamma$ (or $b$ equivalently). Different kernel geometries can thus fit every training label exactly while producing different limiting test errors.

\paragraph{The kernel angle determines whether early stopping is needed.} The location of the minimum generalization error along the trajectory is governed by a sharp threshold in the kernel geometry. Define \begin{equation} 
\begin{aligned} 
c(\varepsilon)&=\left[\frac{\pi}{2(1-\varepsilon)^2}-1\right]^{-\frac{1}{2}},\quad \gamma_{\rm c}(\varepsilon)=\arctan c(\varepsilon) 
\end{aligned} 
\label{eq:bc} 
\end{equation} 
At zero ridge, $\lambda=0$, for every sample ratio $\alpha\in(0,\infty)$, every fixed finite initialization scale $\delta_0$ and every random-label fraction $0<\varepsilon<1$, the generalization error in Eq.~\eqref{eq:errors} decreases until convergence when $\gamma\le\gamma_{\rm c}(\varepsilon)$. Its minimum is therefore attained at the interpolating endpoint, corresponding to $\tes=\infty$. When $\gamma>\gamma_{\rm c}(\varepsilon)$, the generalization error eventually increases, and its minimum is attained at a finite positive time, $\tes<\infty$. Because zero-ridge gradient flow converges to exact interpolation for every kernel considered here, this is precisely a boundary between kernels for which training to interpolation is optimal and kernels for which early stopping strictly improves upon the interpolating solution. The boundary depends only on the random-label fraction $\varepsilon$.

More generally, for fixed explicit regularization $\lambda\ge0$, 
\begin{equation} 
\tes(\lambda)<\infty\quad\Longleftrightarrow\quad \mu_1>c(\varepsilon)\sqrt{\mu_\star^2+\lambda}. \label{eq:boundary} 
\end{equation} 
In Appendix~\ref{app:proof} we establish this result by analyzing the derivative of the ratio $m_\tau^2/q_\tau$ in Eq.~\eqref{eq:mq}. Remarkably, the boundary is independent of the sample ratio~$\alpha$ and the initialization scale $\delta_0$. The same criterion extends to input covariances with bounded eigenvalues when the ground-truth direction is independent of the covariance eigenbasis and isotropically distributed; precise assumptions are given in Appendix~\ref{app:covariance}. It is also unchanged under Euclidean gradient flow in the dual coefficients, as shown in Appendix~\ref{app:metric}.

\begin{figure}[!t] \centering \includegraphics[width=0.6\linewidth]{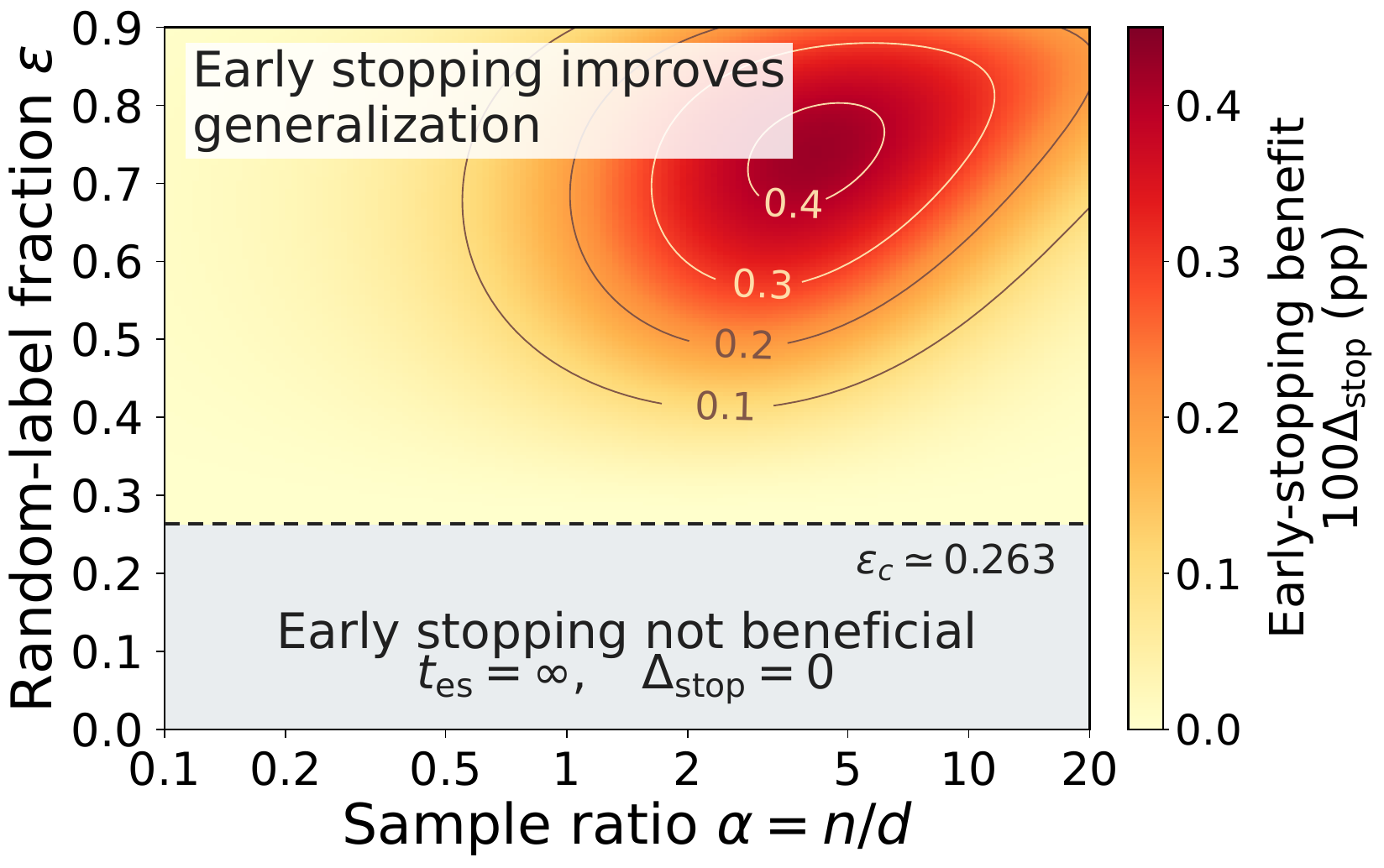} \caption{\textbf{The generalization cost of training to interpolation.} For $\gamma=\pi/5$, $\delta_0=0$, and $\lambda=0$, colors show $100\Delta_{\rm stop}$ from Eq.~\eqref{eq:stopping-benefit-map} in percentage points, comparing interpolation with optimal stopping for the same kernel. The gray region has exactly zero stopping benefit. The dashed line, $\varepsilon_{\rm c}\simeq0.2633$ (for $\gamma=\pi/5$), separates it from a region with a finite optimal stopping time, independently of $\alpha$. Contours mark benefits of $0.1$, $0.2$, $0.3$, and $0.4$ percentage points. } \label{fig:phase-suboptimality} \end{figure}

\paragraph{The cost of training to interpolation.} The boundary above determines whether early stopping improves generalization. To quantify the size of this improvement, we compare the interpolating endpoint with the best point along the trajectory of the same kernel at fixed $\gamma$: \begin{equation} 
\Delta_{\rm stop}(\alpha,\varepsilon;\gamma) =\Egen(\infty;\gamma)-\min_{\tau\in[0,\infty]}\Egen(\tau;\gamma) \label{eq:stopping-benefit-map} 
\end{equation} 
A positive value is the test-error penalty incurred by continuing to interpolation; a zero value means that convergence is optimal for the chosen kernel. Figure~\ref{fig:phase-suboptimality} shows this quantity as the sample ratio $\alpha$ and the fraction of random labels $\varepsilon$ vary, at fixed $\gamma=\pi/5$, zero initialization, and $\lambda=0$. For this kernel, the condition $\gamma=\gamma_{\rm c}(\varepsilon)$ gives $\varepsilon_{\rm c} =1-\sqrt{\frac{\pi}{2}}\sin\gamma \simeq 0.2633$. Below and on this line, $\Delta_{\rm stop}=0$; above it, finite-time stopping strictly improves on interpolation. The sample ratio does not move the boundary, but it changes the cost of training beyond the optimum. The phase diagram therefore distinguishes the existence of an early-stopping benefit from its magnitude. Here optimality is assessed at a fixed kernel; the comparison across kernel geometries follows in the next section.

\section{Optimal interpolation at the critical geometry}

\label{sec:stopping}

The preceding section determined, for each fixed kernel geometry, represented by $\gamma$, whether generalization is optimized by early stopping or by training to interpolation. We now set $\lambda=0$ and ask the stronger question of which kernel geometry and training time jointly attain the lowest generalization error. We will show that this happens for the critical kernel family \begin{equation} \mu_1=c(\varepsilon)\mu_\star\quad\Longleftrightarrow\quad \gamma=\gamma_{\rm c}(\varepsilon). 
\label{eq:angle} 
\end{equation}

Applying Cauchy--Schwarz to Eq.~\eqref{eq:mq} gives \begin{equation} 
\frac{m_\tau^2}{q_\tau}\le I_\star :=\alpha\left\langle\frac{s}{s+c(\varepsilon)^{-2}}\right\rangle_\rho,\qquad  \Egen(\tau)\ge\frac1\pi\arccos\sqrt{I_\star}.
\label{eq:global-optimum}
\end{equation} 

This bound holds at every finite training time for every smooth dot-product kernel satisfying our assumptions. At the critical geometry, i.e. for kernels at $\gamma =\gamma_c(\varepsilon)$ from Eq.~\eqref{eq:bc}, equality is attained at convergence where $F_\infty(s)=s/[s+c(\varepsilon)^{-2}]$. Consequently, the main result of the paper can be stated as 
\begin{equation} 
\Egen(\infty;\gamma_{\rm c}(\varepsilon)) = \inf_{\gamma,\,\tau\geq0}\Egen(\tau;\gamma). 
\end{equation}

Kernels from the critical kernel family with $\gamma=\gamma_{\rm c}(\varepsilon)$ therefore interpolate every training label while attaining the lowest generalization error available across the entire kernel family and all training times. In particular, no finite-time iterate for another unregularized kernel can match or improve upon this interpolating endpoint. This is stronger than benign interpolation: the interpolating predictor is not merely competitive, but minimizes the generalization error over all kernel geometries and all earlier stopping times considered here.

The same critical relation as in Eq.~\eqref{eq:angle} was identified to be relevant for the ERM solutions by \citet{farne2026}; the new dynamical conclusion is that this endpoint also outperforms every early-stopped solution. Appendix~\ref{app:equilibrium} gives the equality conditions, and Figure~\ref{fig:angle-optimality} compares converged and optimally stopped errors across kernel angles and initialization scales.

\begingroup \paragraph{When is the bound attained?} Equality in Eq.~\eqref{eq:global-optimum} requires the initialization contribution to $q_\tau$ in Eq.~\eqref{eq:mq} to vanish and 
\begin{equation} 
F_\tau(s)=C\,\frac{s}{s+b_{\rm c}},\quad b_{\rm c}=c(\varepsilon)^{-2}={\rm cot}^2\gamma_c(\varepsilon), \label{eq:main-equality-condition} 
\end{equation} 
for some constant $C>0$ and on the positive spectral support. The critical angle $\gamma_{\rm c}(\varepsilon)$ is the unique angle attaining this bound, and it does so only at convergence. In Figure~\ref{fig:angle-optimality}, we plot converged and best stopped test errors against $\gamma$. The dashed curve for $\delta_0=0$ looks nearly flat above $\gamma_{\rm c}(\varepsilon)$, but remains strictly above the critical value as shown in the inset. The gap is more visible for $\delta_0=1$. In Appendix~\ref{app:equilibrium} we provide the equality argument.

\begin{figure}[!htbp] \centering \includegraphics[width=0.6\linewidth]{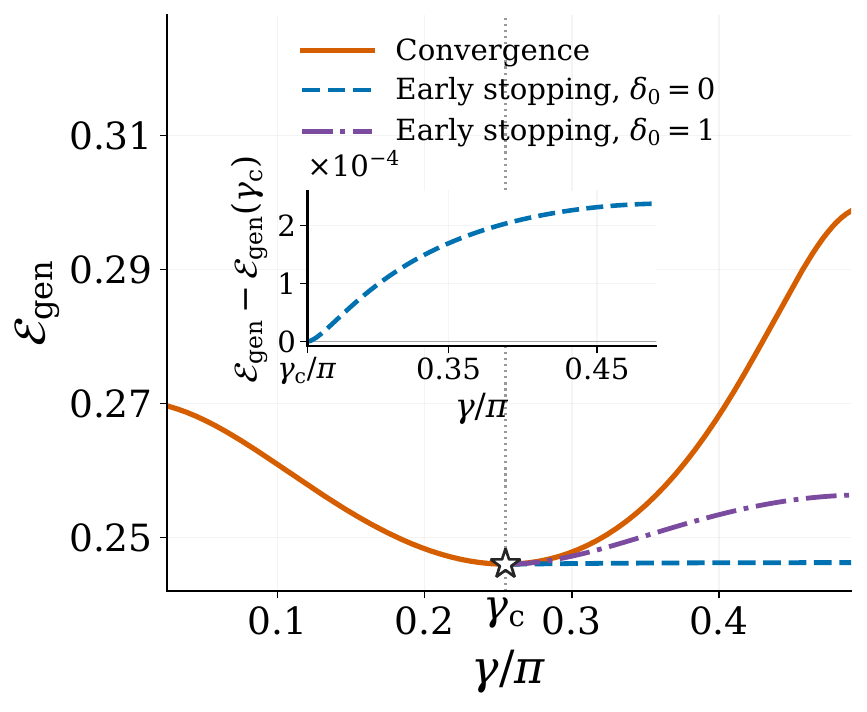} \caption{\textbf{Test error at interpolation and optimal stopping.} We fix $\alpha=1.5$, $\varepsilon=0.1$, and $\lambda=0$ and evaluate Eqs.~\eqref{eq:errors} and~\eqref{eq:mq}. The solid curve is $\Egen(\infty;\gamma)$; the dashed and dash-dotted curves are $\Egen(\taue;\gamma)$ for $\delta_0=0$ and $1$. For $\gamma\le\gamma_{\rm c}(\varepsilon)$, Eq.~\eqref{eq:boundary} gives $\tes=\infty$, so all three curves coincide. The star marks the unique global optimum at $\gamma_{\rm c}(\varepsilon)/\pi\simeq0.2550$ from Eq.~\eqref{eq:bc}. For larger angles, finite optimal stopping remains strictly above this value; the gap is more visible for $\delta_0=1$. The inset zooms on the increasing excess $\Egen(\taue;\gamma)-\Egen(\infty;\gamma_{\rm c}(\varepsilon))$ for the dashed blue curve at $\delta_0=0$, to highlight that the minimum is truly attained only at $\gamma=\gamma_c(\varepsilon)$. } \label{fig:angle-optimality} \end{figure} \endgroup

\begin{figure}[!htbp] \centering \includegraphics[width=0.6\linewidth]{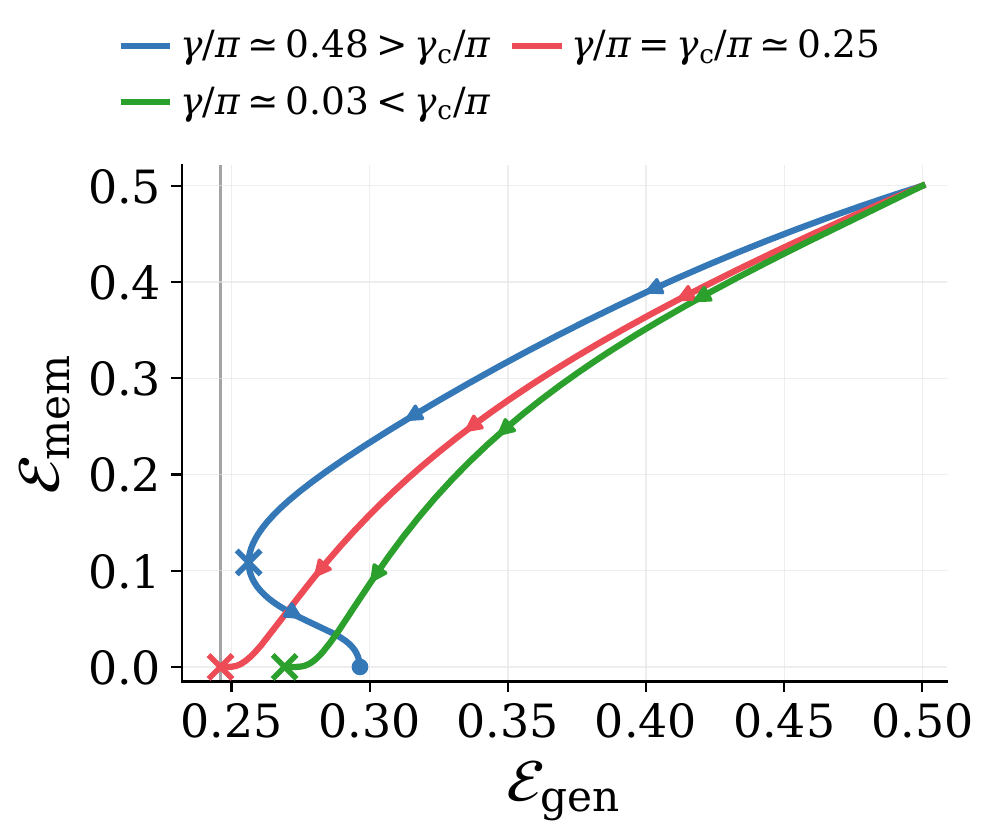} \caption{\textbf{Kernel geometry controls the gradient-flow trajectory.} We fix $\alpha=1.5$, $\varepsilon=0.1$, $\delta_0=1$, and $\lambda=0$. Arrows indicate increasing time; the cross marks test-error minima and the blue dot marks convergence. The blue trajectory benefits from early stopping. Red and green improve until interpolation, but only the critical red kernel attains the test-error bound in Eq.~\eqref{eq:global-optimum} (vertical line).} \label{fig:theory-parametric} \end{figure}

\paragraph{Three geometries, three generalization--memorization paths.} Figure~\ref{fig:theory-parametric} displays the analytical trajectories in the same coordinates as the neural-network trajectories in Figure~\ref{fig:capacity}. The data distribution, sample ratio, and initialization are fixed, while we vary the kernel angle $\gamma$. We distinguish three regimes, corresponding to the following kernel families.

\begin{itemize} \item {\bf Rule-dominated kernels,} $\gamma > \gamma_{\rm c}(\varepsilon)$, generalization initially improves but eventually deteriorates as the randomly labelled examples are fitted, producing a finite optimal stopping time.

\item {\bf Fact-dominated kernels,} $\gamma< \gamma_{\rm c}(\varepsilon)$, the generalization error continues to decrease until interpolation, but the resulting predictor remains suboptimal relative to the critical kernel.

\item {\bf Optimal kernels,} $\gamma = \gamma_{\rm c}(\varepsilon)$: the trajectory reaches the lowest (among all considered kernels and stopping times) test error at the interpolating endpoint, where the memorization error vanishes. Thus, eliminating the benefit of early stopping is not sufficient: optimality is attained specifically at the boundary between the finite-time and interpolation-optimal regimes. \end{itemize}

All three kernel families have enough capacity to eventually interpolate the training data. What differs is how fitting the unpredictable labels affects predictions on new examples. The critical geometry provides a fixed-representation mechanism through which interpolation can be completed without sacrificing the best generalization attainable within the kernel family. Its trajectory qualitatively resembles those of the larger, deeper neural networks in Figure~\ref{fig:capacity}. The kernel analysis does not, however, explain how neural-network training constructs a representation with analogous properties. Determining which aspects of feature learning, model size, and architecture produce this behavior remains an open question.

\begin{figure}[!htbp] \centering \includegraphics[width=0.6\linewidth]{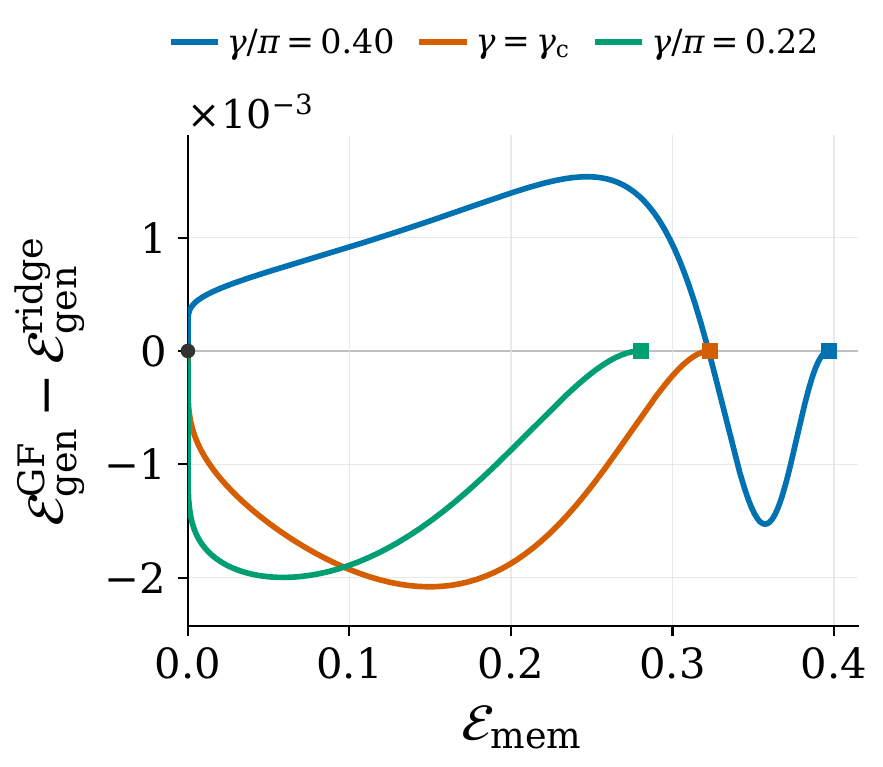} \caption{\textbf{Gradient flow and ridge follow different parametric paths.} We fix $\alpha=5$, $\varepsilon=0.1$, and $\delta_0=0$, with $\gamma/\pi=0.40$, $\gamma=\gamma_{\rm c}(\varepsilon)$ ($\gamma_{\rm c}(\varepsilon)/\pi\simeq0.2550$), and $\gamma/\pi=0.22$. For each kernel, we plot the test-error difference between unregularized kernel gradient flow and converged kernel ridge regression at the same memorization error. Negative values favor gradient flow. Squares mark the shared $t\to0^+$ and $\lambda\to\infty$ classification limits; the circle marks the shared interpolating endpoint. Curves are numerical evaluations of Eqs.~\eqref{eq:errors}--\eqref{eq:VQ}; we use the ERM ridge result of \citet{farne2026}. More details in Appendix~\ref{app:numerical-checks}.} \label{fig:gf-ridge-difference} \end{figure}

\paragraph{Early stopping and ridge follow different paths.}

Early stopping and ridge regression both regularize least squares, but generally give different predictors \citep{yao2007,ali2019,stark2025}. Here we compare their generalization and memorization errors in the RAF model, and relate the optimal stopping time to the optimal ridge parameter found by \citet{farne2026}. We set $\delta_0=0$. From Eq.~\eqref{eq:filters}, the spectral filters $F_{\tau,\ell}$, with reduced time $\tau=\mu_1^2t$ and reduced ridge regularization $\ell=\lambda/\mu_1^2$, for unregularized gradient flow and converged ridge regression are 
\begin{equation} 
\begin{aligned} 
F_{\tau,0}(s)&=\frac{s\bigl(1-e^{-\tau(s+b)}\bigr)}{s+b},\\ F_{\infty,\ell}(s)&=\frac{s}{s+b+\ell},\qquad \ell=\frac{\lambda}{\mu_1^2}. 
\end{aligned} 
\label{eq:time-ridge-filters} 
\end{equation} 
with $b=\cot^2\gamma$. The first expression has $\lambda=0$; the second is evaluated after training to convergence. To make them equal at a positive eigenvalue $s$, solve $F_{\tau,0}(s)=F_{\infty,\ell}(s)$ for $\ell$. This gives 
\begin{equation} 
\ell_\tau(s)=\frac{s+b}{e^{\tau(s+b)}-1}. 
\label{eq:mode-dependent-ridge} 
\end{equation} 
The required value depends on $s$. This is the generalized-ridge interpretation of gradient flow \citep{ali2019}.

For a fixed kernel angle, the optimal nonnegative ridge regularization is \citep{farne2026} 
\begin{equation} 
\lambda_{\rm opt}=\mu_1^2\max\{\cot^2\gamma_{\rm c}(\varepsilon)-\cot^2\gamma,0\}. 
\label{eq:optimal-ridge} 
\end{equation}

When $\gamma>\gamma_{\rm c}(\varepsilon)$, this choice is positive and makes $F_{\infty,\lambda_{\rm opt}/\mu_1^2}(s)=s/(s+\cot^2\gamma_{\rm c}(\varepsilon))$, exactly the filter of the unregularized critical kernel at convergence. Ridge therefore attains the bound in Eq.~\eqref{eq:global-optimum} while retaining positive training square loss. Finite-time gradient flow does not satisfy the equality condition in Eq.~\eqref{eq:main-equality-condition}, so \begin{equation} 
\Egen(\taue;\gamma,\lambda=0) >\Egen(\infty;\gamma,\lambda=\lambda_{\rm opt})=\Egen(\infty;\gamma_{\rm c}(\varepsilon),\lambda=0), \qquad \gamma>\gamma_{\rm c}(\varepsilon). \label{eq:stopping-ridge-gap} 
\end{equation}

For $\gamma\le\gamma_{\rm c}(\varepsilon)$, both procedures attain their lowest test error at the shared interpolating endpoint, $\tes=\infty$ and $\lambda_{\rm opt}=0$; among these kernels, only $\gamma=\gamma_{\rm c}(\varepsilon)$ attains the bound.

The distinction also concerns the entire generalization-memorization path. Figure~\ref{fig:gf-ridge-difference} compares the two test errors at the same memorization error, making their small differences clearly visible.

\section{Discussion and Conclusion}

In this work, we asked whether the presence of extensive label noise necessarily makes early stopping preferable to interpolation. The noisy CIFAR10 experiments in Fig.~\ref{fig:capacity} and Appendix~\ref{app:nn} support the hypothesis that deep and highly overparameterized neural networks actually do not benefit from early stopping. Figure~\ref{fig:capacity} shows that the larger and deeper networks attain their best observed test performance only after their memorization error reaches zero, while fixed-feature and shallower models exhibit late-time test-error deterioration. Appendix~\ref{app:nn} shows that this observation is robust beyond the setting of Fig.~\ref{fig:capacity}: it persists across a range of training recipes, including cross-entropy, data augmentation, and a scheduled learning rate, as well as across additional optimization controls.

We then asked whether the absence of a benefit from early stopping can be reproduced in an analytically solvable setting. Our analysis of high-dimensional learning curves for kernel gradient flow shows that it can. Kernel geometry determines whether the lowest test error occurs at a finite time or at convergence. We identify a critical kernel geometry for which the interpolating predictor attains the lowest test error across the considered kernel family and all training times. Thus, fitting unpredictable labels need not come at the expense of generalization, in a sense yet considerably stronger than the one known from the literature on benign overfitting.

The critical-kernel trajectory provides a concrete fixed-representation counterpart of the behavior observed in the deeper neural networks: fitting the unpredictable labels does not produce a late-time deterioration of generalization, and the best test error is attained at the end of the trajectory. Our analysis does not, however, establish that the networks learn this same geometry, nor does it explain how feature learning, architecture, and optimization produce the empirically observed behavior. The theory nevertheless places the empirical hypothesis formulated in the introduction on a precise footing: even under extensive label noise, the absence of a benefit from early stopping is possible and can be controlled by representation geometry. Understanding how larger and deeper feature-learning architectures realize analogous behavior remains an open question.

All the results can be reproduced using the code provided in the repository: \url{https://github.com/SPOC-group/Early-stopping}.

\subsection*{Acknowledgement}

We acknowledge funding from the Swiss National Science Foundation grants SNSF SMArtNet (grant number 212049), the Simons Collaboration on the Physics of Learning and Neural Computation via the Simons Foundation grant ($\#$1257413 (LZ)), and the Swiss State Secretariat for Education, Research and Innovation (SERI), which funded the ERC Advanced Grant SPASM-net.

We thank Giulio Biroli for his question about extending the work \cite{farne2026} to the dynamical domain, and Gabriele Farné for useful discussions.

\subsection*{AI Use Statement} In this work, we used generative AI tools to help check, and polish mathematical derivations, as well as for literature searches. Generative AI was also used to provide feedback on numerical methodology, in particular data preprocessing and code development. We also used AI to polish the writing. We have reviewed all AI-assisted work and we take responsibility for the final content of this work, including text, claims or artifacts produced with the aid of generative AI.

\begingroup \small \setlength{\bibsep}{4pt plus 1pt} \bibliography{biblio} \endgroup \bibliographystyle{plainnat}

\appendix \onecolumn

\section{From a nonlinear kernel to an effective linear predictor} \label{app:kernel}

In this appendix we recall the model and learning problem, then explain why a nonlinear kernel has a simple high-dimensional description. We give the full kernel gradient-flow solution, calculate the RAF label moments, and then justify the effective linear predictor and removal of the constant kernel component. Appendix~\ref{app:derivations} uses this description to derive the four order parameters and the the low dimensional description for the generalization and memorization errors.

\paragraph{Setting and notation.} We work in a supervised setting. We start from $n$ inputs $x^\mu\in\R^d$ that are independent and uniform on the sphere $\|x^\mu\|_2=\sqrt d$. A vector $w_\star\sim\N(0,\Id_d)$ specifies a common rule to be learned and is independent of the inputs. With probability $1-\varepsilon$, sample $\mu$ has label $y^\mu=\sign(g^\mu)$, where $g^\mu=x^{\mu\top}w_\star/\sqrt d$; with probability $\varepsilon$, it has an independent, equally likely label $y^\mu=\pm1$. We call the latter samples facts and denote their indices by $\mathcal I_{\rm mem}$. The rows of $X\in\R^{n\times d}$ are $x^{\mu\top}$, and $y\in\R^n$ collects the labels. In particular, $\E y^\mu=0$ and $\E[(y^\mu)^2]=1$.

We learn this data model with a fixed positive-semidefinite dot-product kernel $k(x,x')=\sigma(x^\top x'/d)$, with training matrix $K_{\mu\nu}=k(x^\mu,x^\nu)$ and test vector $k(x)_\mu=k(x^\mu,x)$. We assume the function $\sigma$ has three continuous derivatives near zero and is finite at one. We define the coefficients 
\begin{equation} 
\mu_0^2=\sigma(0)\ge0,\qquad \mu_1^2=\sigma'(0)>0,\qquad \mu_\star^2=\sigma(1)-\sigma(0)-\sigma'(0)>0. 
\label{eq:appendix-kernel-coefficients} 
\end{equation} 

The kernel angle is defined as $\gamma=\arctan(\mu_1/\mu_\star)\in(0,\pi/2)$. For example, a power series $\sigma(r)=\sum_{j\ge0}c_jr^j$ with nonnegative summable coefficients, $c_1>0$, $\sum_{j\ge2}c_j>0$ and the stated smoothness defines an admissible kernel.

Writing $\mathcal H_k$ for its reproducing-kernel Hilbert space, kernel ridge regression minimizes the regularized square loss: 

\begin{equation} 
\mathcal L_\lambda(f)=\frac12\sum_{\mu=1}^n[y^\mu-f(x^\mu)]^2 +\frac\lambda2\|f\|_{\mathcal H_k}^2. 
\label{eq:functional-risk} 
\end{equation} 

We follow the RKHS gradient flow of this functional, starting from $f_0(\cdot)=\sum_\mu a_{0,\mu}k(x^\mu,\cdot)$ with independent $\bm a_0\sim\N(0,\Delta_0\Id_n)$. The predictor remains in this span, because its time derivative is a linear combination of the same training kernel functions, namely $f_t(x)=k(x)^\top\bm a_t$. The loss is summed over samples, rather than divided by $n$. Generalization is the probability of disagreeing with the rule on a new independent input, while memorization is the classification error on the training facts, as in Eqs.~\eqref{eq:gen-def} and~\eqref{eq:mem-def} of the main text.

Throughout these two appendices, the high-dimensional limit we consider is $n,d\to\infty$ at fixed $n/d\to\alpha\in(0,\infty)$, with $0<\varepsilon<1$, a fixed kernel, fixed $\lambda\ge0$, finite $\Delta_0$, and fixed $t>0$. The time limit $t\to\infty$ is taken afterwards. Unless otherwise stated, a vanishing random error means convergence in probability in the high-dimensional limit: its probability of exceeding any fixed positive tolerance tends to zero. The spherical inputs have population covariance $\Id_d$; general covariance is treated separately in Appendix~\ref{app:covariance}.

\paragraph{Kernel matrix.} Write the sample covariance $C_n=XX^\top/d$. Its diagonal entries equal one, and off-diagonal entries are of order $d^{-1/2}$. Its largest eigenvalue remains bounded at fixed $n/d$. Here $C_{n,\mu\nu}=x^{\mu\top}x^\nu/d$, so the kernel entries are $K_{\mu\nu}=\sigma(C_{n,\mu\nu})$. Expanding $\sigma(\cdot)$ around zero therefore gives a constant and a linear term; the quadratic term has off-diagonal mean $\sigma''(0)/(2d)$. The diagonal is evaluated directly at $\sigma(1)$. The resulting kernel expansion is taken from \citet{elkaroui2010}, with related extensions in \citet{dovu2013}, and reads 
\begin{equation} K=c_d\bm1\bm1^\top+\mu_1^2C_n+\mu_\star^2\Id_n+R_d, \qquad c_d=\mu_0^2+\frac{\sigma''(0)}{2d}, \qquad \|R_d\|_{\op}\longrightarrow0,
\label{eq:full-kernel-reduction} 
\end{equation} 
where $\|\cdot\|_{\op}$ is the largest singular value and the convergence is in the high-dimensional limit just specified. This is Eq.~\eqref{eq:kernel}. Spherical inputs can be represented as $x^\mu=\sqrt d\,z^\mu/\|z^\mu\|_2$ with independent standard Gaussian $z^\mu$; their normalization factors tend uniformly to one, giving the corresponding spherical form of the kernel expansion.

For Gaussian random-feature kernels, the constant, linear and higher-degree contributions are the squared Hermite coefficients of the underlying activation \citep{mei2022,farne2026}. Throughout this work we assume both $\mu_1^2>0$ and $\mu_\star^2>0$; the latter condition excludes purely linear kernels.

\paragraph{RKHS gradient flow and its solution.} For the functional in Eq.~\eqref{eq:functional-risk}, the reproducing identity $u(x^\mu)=\langle u,k(x^\mu,\cdot)\rangle_{\mathcal H_k}$, which expresses evaluation at a sample as an RKHS inner product, yields \begin{equation} 
\nabla_{\!\mathcal H_k}\mathcal L_\lambda(f) =\sum_\mu[f(x^\mu)-y^\mu]k(x^\mu,\cdot)+\lambda f.
\label{eq:functional-gradient} 
\end{equation} 
The gradient is taken with respect to the function $f$, with the RKHS inner product; the kernel itself remains fixed. Substituting $f_t=\sum_\mu a_{t,\mu}k(x^\mu,\cdot)$ into $\dot f_t=-\nabla_{\!\mathcal H_k}\mathcal L_\lambda(f_t)$ gives the coefficient dynamics and their solution, 
\begin{equation} 
\dot{\bm a}_t=y-(K+\lambda\Id_n)\bm a_t,\qquad \bm a_t=e^{-t(K+\lambda\Id_n)}\bm a_0 +(K+\lambda\Id_n)^{-1}(\Id_n-e^{-t(K+\lambda\Id_n)})y. 
\label{eq:appendix-flow-solution} 
\end{equation} 
This is the standard kernel least-squares solution \citep{ali2019,ghosh2022}. The alternative Euclidean coefficient gradient has an extra factor $K$, as explained in Appendix~\ref{app:metric}.

\paragraph{RAF label moments.} At fixed $X$, the label-dependent part of Eq.~\eqref{eq:appendix-flow-solution} is linear in $y$. Its cross-moment with the rule vector $\E[\bm a_t w_\star^\top\mid X]$ therefore requires $\E[yw_\star^\top\mid X]$, while its covariance $\E[\bm a_t\bm a_t^\top\mid X]$ requires $\E[yy^\top\mid X]$ and the initialization variance. We calculate these label moments first; their role in the prediction errors will become explicit after reducing the predictor below. We separate each label into a part correlated with the rule field $g=Xw_\star/\sqrt d$ and a residual, as in the square-loss RAF and Gaussian-covariate calculations \citep{farne2026,loureiro2021}. Conditional on $X$, $g=Xw_\star /\sqrt{d}$ is Gaussian with covariance $C_n$ and unit marginal variance since $\E[gg^\top\mid X]=C_n$ and $C_{n,\mu\mu}=\|x^\mu\|_2^2/d=1$. For $G\sim\N(0,1)$, $\E[G\sign G]=\E|G|=\sqrt{2/\pi}$; the independent random labels contribute zero correlation because each factual label is independent of $g^\mu$ and has mean zero. Hence 
\begin{equation} 
a=\E[y^\mu g^\mu]=(1-\varepsilon)\sqrt{2/\pi},\qquad y=ag+\xi,\qquad \xi:=y-ag. 
\label{eq:signal-residual} 
\end{equation} 
This is an exact decomposition of the binary labels. To see what moments it supplies, we use the Gaussian identities $\E[G_1\sign G_2]=\sqrt{2/\pi}\,\E[G_1G_2]$ and $\E[\sign G_1\sign G_2]=(2/\pi)\arcsin(\E[G_1G_2])$ for unit-variance $G_1,G_2$. They give 
\begin{equation} 
\E[yw_\star^\top\mid X]=\frac{aX}{\sqrt d},\qquad H_{\mu\nu}:=\E[y^\mu y^\nu\mid X] =
\begin{cases}1,&\mu=\nu,\\a^2\arcsin(C_{n,\mu\nu}),&\mu\ne\nu. 
\end{cases} 
\label{eq:label-moments} 
\end{equation}

Since $g=Xw_\star/\sqrt d$, Eq.~\eqref{eq:label-moments} also gives $\E[gw_\star^\top\mid X]=X/\sqrt d$ and $\E[yg^\top\mid X]=aC_n$. Expanding $\arcsin r=r+O(r^3)$ at the small off-diagonal overlaps and then expanding the products with $\xi=y-ag$ gives
\begin{equation}
\begin{aligned}
H&=a^2C_n+(1-a^2)\Id_n+o(1),\\
\E[\xi w_\star^\top\mid X]
&=\frac{aX}{\sqrt d}-a\frac{X}{\sqrt d}=0,\\
\E[\xi\xi^\top\mid X]
&=H-a^2C_n-a^2C_n+a^2C_n\\
&=H-a^2C_n=(1-a^2)\Id_n+o(1).
\end{aligned}
\label{eq:label-covariance}
\end{equation}

The term $a^2C_n$ has diagonal $a^2$; adding $(1-a^2)\Id_n$ restores $H_{\mu\mu}=1$. Here and below, $o(1)$ denotes a correction vanishing in probability in the proportional high-dimensional limit. The decomposition $y=ag+\xi$ and the zero cross-moment $\E[\xi w_\star^\top\mid X]=0$ are exact at finite dimension. The covariance simplification is asymptotically exact: the omitted off-diagonal entries are $O(C_{n,\mu\nu}^3)$, whose squared sum is $O_{\mathbb P}(n^2/d^3)=O_{\mathbb P}(d^{-1})$. Their Frobenius norm, and hence their operator norm, therefore vanishes. The labels are centered conditional on $X$, by the symmetry of the Gaussian rule field and the independent random signs. The residual $\xi$ need not be Gaussian or have independent entries. Only its covariance and cross-moment enter the overlap and squared-norm calculations below; the Gaussian limit of factual margins is justified separately.

\paragraph{Prediction on an independent test input.} For a new input $x$, the test vector is $k(x)_\mu=\sigma(x^{\mu\top}x/d)$ and the prediction is $f_t(x)=\sum_\mu a_{t,\mu}k(x)_\mu$. All these overlaps are small: unlike the training matrix, the test vector has no diagonal entry $k(x^\mu,x^\mu)=\sigma(1)$. Expanding each entry of the test vector, $k(x)_\mu=\sigma(x^{\mu\top}x/d)$, gives 
\begin{equation} 
k(x)_\mu=c_d+\frac{\mu_1^2}{d}x^{\mu\top}x+\zeta_\mu(x),\qquad f_t(x)=c_d\bm1^\top\bm a_t +\frac{\mu_1^2}{d}x^\top X^\top\bm a_t +\zeta(x)^\top\bm a_t. 
\label{eq:test-expansion} 
\end{equation} 
Here $\zeta_\mu$ contains the quadratic term with its mean removed, $\zeta_\mu^{(2)}=\frac12\sigma''(0)[(x^{\mu\top}x/d)^2-1/d]$, and the higher-order terms. The spherical fourth-moment identity $\E_x[x_ix_jx_kx_l]=d(\delta_{ij}\delta_{kl}+\delta_{ik}\delta_{jl}+\delta_{il}\delta_{jk})/(d+2)$, together with $C_{n,\mu\nu}=x^{\mu\top}x^\nu/d$, gives, conditional on the training inputs, 
\begin{equation} 
\E_x[\zeta_\mu^{(2)}\zeta_\nu^{(2)}\mid X] =\frac{\sigma''(0)^2}{2d(d+2)}\left(C_{n,\mu\nu}^2-\frac1d\right). 
\label{eq:appendix-test-remainder} 
\end{equation} 
To control the sum over all training examples, first note that the coefficients have total squared norm of order $n$ at fixed time. For an eigenvalue $\Lambda_j\ge0$ of $K+\lambda\Id_n$, Eq.~\eqref{eq:appendix-flow-solution} has label response $\int_0^t e^{-s\Lambda_j}\,ds\le t$. Since $\|y\|_2^2=n$ and the initialization is independent and centered,
\begin{equation}
\E\|\bm a_t\|_2^2\le n(\Delta_0+t^2).
\label{eq:coefficient-energy}
\end{equation}
Now write $R_2(x)=\sum_\mu a_{t,\mu}\zeta_\mu^{(2)}(x)$ and average over the independent test input while holding $X$ and $\bm a_t$ fixed. Equation~\eqref{eq:appendix-test-remainder} gives
\begin{equation}
\begin{aligned}
\E_x[R_2(x)^2\mid X,\bm a_t]
&=\frac{\sigma''(0)^2}{2d(d+2)}\bigg[\left(1-\frac1d\right)\sum_\mu a_{t,\mu}^2\\
&\qquad+\sum_{\mu\ne\nu}a_{t,\mu}a_{t,\nu}\left(C_{n,\mu\nu}^2-\frac1d\right)\bigg].
\end{aligned}
\label{eq:test-remainder-bound}
\end{equation}
The diagonal sum is $O(n)$ by Eq.~\eqref{eq:coefficient-energy}. For the off-diagonal sum, the spherical moments $\E C_{n,\mu\nu}^2=1/d$ and $\E C_{n,\mu\nu}^4=3/[d(d+2)]$ give
\begin{equation}
D_X:=\sum_{\mu\ne\nu}\left(C_{n,\mu\nu}^2-\frac1d\right)^2,
\qquad \E D_X=\frac{2n(n-1)(d-1)}{d^2(d+2)}=O(1).
\end{equation}
Cauchy--Schwarz therefore bounds the absolute off-diagonal sum by $(\sum_\mu a_{t,\mu}^2)\sqrt{D_X}=O(n)$. Both terms in brackets are of order $n$, and the mean squared correction is $O(n/d^2)\to0$. This calculation does not assume independence between the coefficients and training inputs. Higher Taylor terms involve at least three overlaps; their sixth moments are $O(d^{-3})$, giving a total correction of order $n^2/d^3\to0$. Combining these vanishing terms with Eq.~\eqref{eq:test-expansion} leaves only the constant offset and linear prediction term.

Thus the full prediction has the form
\begin{equation}
f_t(x)=c_d\bm1^\top\bm a_t+\frac{\mu_1^2}{d}x^\top X^\top\bm a_t+o(1).
\label{eq:prediction-before-constant-removal}
\end{equation}
We now show that the constant offset and its effect on the remaining coefficients also vanish.

\paragraph{Why the constant does not affect the limiting errors.} Equation~\eqref{eq:prediction-before-constant-removal} separates the prediction into a constant and an input-dependent part. We denote the constant by $b_t=c_d\bm1^\top\bm a_t$, recalling that $c_d=\mu_0^2+\sigma''(0)/(2d)$. We now show that $b_t$ vanishes at every fixed positive time. Substituting the kernel expansion in Eq.~\eqref{eq:full-kernel-reduction} into the coefficient flow, and omitting its vanishing remainder, gives
\begin{equation}
\dot{\bm a}_t=y-\mu_1^2C_n\bm a_t-(\mu_\star^2+\lambda)\bm a_t-b_t\bm1.
\end{equation}
Multiply this equation by $c_d\bm1^\top$. Since $c_d\bm1^\top\bm a_t=b_t$ and $\bm1^\top\bm1=n$, the last two terms become $-(\mu_\star^2+\lambda)b_t$ and $-nc_db_t$. Therefore
\begin{equation}
\dot b_t+(nc_d+\mu_\star^2+\lambda)b_t
=c_d\left(\bm1^\top y-\mu_1^2\bm1^\top C_n\bm a_t\right).
\label{eq:constant-mode-solution}
\end{equation}
We calculate the mean squared offset, averaging over the rule vector, label assignments and initialization. We first keep $X$ fixed so that the matrices governing the flow are known. The averages then use the centered label covariance $H=\E[yy^\top\mid X]$ derived above. Its bounded eigenvalues give
\begin{equation}
\E[(\bm1^\top y)^2\mid X]
=\bm1^\top H\bm1
\le n\lambda_{\max}(H)=O(n).
\end{equation}
Here $\lambda_{\max}$ denotes the largest eigenvalue.
To estimate the second sum, we use the full matrix governing the coefficient dynamics,
\begin{equation}
M=\mu_1^2C_n+(\mu_\star^2+\lambda)\Id_n+c_d\bm1\bm1^\top.
\end{equation}
At an intermediate time $0\le r\le t$, the solution is
\begin{equation}
\bm a_r=e^{-rM}\bm a_0+R_r y,
\qquad
R_r=\int_0^r e^{-uM}\,du.
\end{equation}
For an eigenvalue $\Lambda\ge0$ of $M$, the initialization is multiplied by $e^{-r\Lambda}\le1$, while the labels are multiplied by $\int_0^r e^{-u\Lambda}\,du\le r$. These are the two factors controlling the fluctuations of $\bm a_r$. More explicitly, for any vector $p$ fixed once $X$ is fixed, independence of the centered initialization and the labels gives
\begin{equation}
\begin{aligned}
\E[(p^\top\bm a_r)^2\mid X]
&=\Delta_0\|e^{-rM}p\|_2^2
 +(R_rp)^\top H(R_rp)\\
&\le
\left[\Delta_0+r^2\lambda_{\max}(H)\right]\|p\|_2^2.
\end{aligned}
\end{equation}
The sum needed here is obtained by choosing $p=C_n\bm1$. Thus $C_n\bm1$ supplies the weights of the sum, while $M$ determines their time evolution. We obtain
\begin{equation}
\begin{aligned}
\E[(\bm1^\top C_n\bm a_r)^2\mid X]
&\le
\left[\Delta_0+r^2\lambda_{\max}(H)\right]
\bm1^\top C_n^2\bm1\\
&\le
n\left[\Delta_0+t^2\lambda_{\max}(H)\right]
\lambda_{\max}(C_n)^2
=O(n).
\end{aligned}
\end{equation}
Both sums on the right-hand side of Eq.~\eqref{eq:constant-mode-solution} thus have fluctuations of order $\sqrt n$. This conclusion does not require them to be independent. The estimates hold for typical input matrices, for which $\lambda_{\max}(H)$ and $\lambda_{\max}(C_n)$ remain bounded.

We first consider $\mu_0^2>0$, so $c_d\to\mu_0^2>0$. Multiplying the scalar equation by $e^{(nc_d+\mu_\star^2+\lambda)t}$ and integrating gives
\begin{equation}
\begin{aligned}
b_t
&=e^{-(nc_d+\mu_\star^2+\lambda)t}b_0\\
&\quad+c_d\int_0^t
e^{-(nc_d+\mu_\star^2+\lambda)(t-r)}
\left(\bm1^\top y-\mu_1^2\bm1^\top C_n\bm a_r\right)\,dr.
\end{aligned}
\end{equation}
The first term comes from initialization. Since the $n$ initial coefficients are independent, centered and have variance $\Delta_0$, we have $\E[b_0^2\mid X]=nc_d^2\Delta_0$. Its contribution therefore decays exponentially at fixed $t>0$.

For the integral term, the preceding variance estimates give a size of order $\sqrt n$ for the expression in parentheses, uniformly over $0\le r\le t$. Here and below, size means root mean square. Integrating these sizes bounds the root mean square of the integral, even though its values at different times are correlated. Its contribution is consequently bounded, up to a factor independent of $n,d$, by
\begin{equation}
c_d\sqrt n\int_0^t
e^{-(nc_d+\mu_\star^2+\lambda)(t-r)}\,dr
=
\frac{c_d\sqrt n\left(1-e^{-(nc_d+\mu_\star^2+\lambda)t}\right)}
{nc_d+\mu_\star^2+\lambda}
\le\frac1{\sqrt n}.
\end{equation}
Thus the initial term disappears exponentially and the integral term has squared size $O(n^{-1})$:
\begin{equation}
\E[b_t^2\mid X]
=
O\!\left(n e^{-2nc_dt}\right)+O(n^{-1})
\longrightarrow0.
\label{eq:constant-offset-limit}
\end{equation}
If $\mu_0=0$, then $c_d=O(d^{-1})$. Applying the coefficient variance estimate with $p=\bm1$ directly gives
\begin{equation}
\E[b_t^2\mid X]
\le c_d^2n\left[\Delta_0+t^2\lambda_{\max}(H)\right]
=O(n/d^2)\longrightarrow0.
\end{equation}
The constant prediction therefore vanishes in both cases.

To justify removing the constant from the training matrix as well, we must compare the coefficients obtained with and without it. Define
\begin{equation}
K_{\rm eff}=\mu_1^2C_n+\mu_\star^2\Id_n.
\end{equation}
Starting from the same initialization and using the same labels, the coefficients for this matrix are explicitly
\begin{equation}
\bm a_t^{\rm eff}
=e^{-t(K_{\rm eff}+\lambda\Id_n)}\bm a_0
+\int_0^t e^{-(t-r)(K_{\rm eff}+\lambda\Id_n)}y\,dr.
\end{equation}
Their equation is $\dot{\bm a}_t^{\rm eff}=y-(K_{\rm eff}+\lambda\Id_n)\bm a_t^{\rm eff}$. Subtracting it from the full coefficient equation and integrating gives
\begin{equation}
\bm a_t-\bm a_t^{\rm eff}
=-\int_0^t e^{-(t-r)(K_{\rm eff}+\lambda\Id_n)}b_r\bm1\,dr.
\end{equation}
Every eigenvalue of $K_{\rm eff}+\lambda\Id_n$ is positive, so the exponential can only decrease the norm of each vector $b_r\bm1$. Since that vector has norm $\sqrt n\,|b_r|$,
\begin{equation}
\left(
\E[\|\bm a_t-\bm a_t^{\rm eff}\|_2^2\mid X]
\right)^{1/2}
\le
\sqrt n\int_0^t\sqrt{\E[b_r^2\mid X]}\,dr
=O(1).
\label{eq:constant-coefficient-change}
\end{equation}
For $\mu_0^2>0$, the initial contribution to this bound is of order $n\int_0^t e^{-nc_dr}\,dr=O(1)$, and the learned contribution is $O(t)$. For $\mu_0=0$, the bound is $O(n/d)$, which also remains finite.
The coefficient difference need not vanish, but its contribution to the errors does. Indeed, the training-prediction difference is $K_{\rm eff}(\bm a_t-\bm a_t^{\rm eff})+b_t\bm1$. The preceding estimates and the bounded eigenvalues of $K_{\rm eff}$ give
\begin{equation}
\frac1n
\E\!\left[
\left\|K_{\rm eff}(\bm a_t-\bm a_t^{\rm eff})+b_t\bm1\right\|_2^2
\middle|X
\right]
=O(n^{-1}).
\end{equation}
For a new input with covariance $\Id_d$, the change in the input-dependent prediction has mean square
\begin{equation}
\E\!\left[
\left(
\frac{\mu_1^2x^\top X^\top(\bm a_t-\bm a_t^{\rm eff})}{d}
\right)^2
\middle|X
\right]
=
\frac{\mu_1^4}{d}
\E\!\left[
(\bm a_t-\bm a_t^{\rm eff})^\top
C_n(\bm a_t-\bm a_t^{\rm eff})
\middle|X
\right]
=O(d^{-1}).
\end{equation}
Here the average also includes the new input $x$. Both prediction differences vanish, as does the constant offset. We may therefore use the effective dynamics in the limiting calculations of $\Egen$ and $\Emem$ below. This argument applies at fixed $t>0$.

In the following calculations we write simply $\bm a_t$ for $\bm a_t^{\rm eff}$: it solves the effective flow displayed above, starting from the original initialization $\bm a_0$. The symbol $f_t(x)$ continues to denote the original prediction, which differs from the effective prediction by a vanishing amount at fixed $t>0$.

After this removal, define the effective weight vector $v_\tau\in\R^d$ by
\begin{equation}
K_{\rm eff}=\mu_1^2C_n+\mu_\star^2\Id_n, \qquad v_\tau=\frac{\mu_1^2X^\top\bm a_t}{\sqrt d}, \qquad f_t(x)-\frac{x^\top v_\tau}{\sqrt d}\longrightarrow0, \qquad \tau=\mu_1^2t.
\label{eq:test-transfer}
\end{equation}
Thus $v_\tau$ is the linear predictor seen by a new input in the high-dimensional limit. The diagonal component $\mu_\star^2\Id_n$ still changes the fitted coefficients $\bm a_t$, but it has no direct diagonal contribution to the test vector. This distinction between the training matrix and test prediction is also central to the static RAF analysis \citep{farne2026}.

\paragraph{Example: the full RBF kernel.} For $\sigma_\eta(r)=e^{-\eta(1-r)}$, Eq.~\eqref{eq:appendix-kernel-coefficients} gives 
\begin{equation} 
\mu_0^2=e^{-\eta},\qquad \mu_1^2=\eta e^{-\eta},\qquad \mu_\star^2=1-(1+\eta)e^{-\eta},\qquad \gamma=\arctan\sqrt{\frac{\eta}{e^\eta-1-\eta}}. 
\label{eq:rbf-masses} 
\end{equation} 
After reducing time by $\mu_1^2$ and fixing the reduced initialization variance, only their ratio enters the unregularized calculation. Thus, the properties of the kernels we consider in our equations are entirely summarized by the kernel angle $\gamma=\arctan(\mu_1/\mu_\star)$. We invert this relation to match a desired angle in numerical checks of the theory. Those simulations use the full RBF kernel, including its nonzero constant term, rather than the effective matrix in Eq.~\eqref{eq:test-transfer}.

\section{Derivation of the order parameters' equations} \label{app:derivations} In this appendix we derive the low-dimensional learning curves from the effective predictor established in the previous Appendix~\ref{app:kernel}. We first solve its reduced eigenmode dynamics, then use the label moments already calculated to evaluate overlaps, prediction variances and factual margins. Finally, normalized traces become one-dimensional spectral integrals. This follows usual square-loss learning-curve calculations \citep{advani2020,loureiro2021,bodin2022}.

To evaluate the solution with the effective kernel $K_{\rm eff}$ previously computed, we define the reduced quantities 
\begin{equation} 
\tau=\mu_1^2t,\qquad b=\frac{\mu_\star^2}{\mu_1^2}=\cot^2\gamma,\qquad \ell=\frac\lambda{\mu_1^2},\qquad \beta=b+\ell,\qquad \widetilde{\bm a}_\tau=\mu_1^2\bm a_t,\qquad \delta_0=\mu_1^4\Delta_0. 
\label{eq:appendix-reduced} 
\end{equation}

Here $b=\cot^2\gamma$ is a shorthand for the kernel geometry. We absorb the factor $\mu_1^2$ into the coefficients by writing $\widetilde{\bm a}_\tau=\mu_1^2\bm a_t$, while retaining $\bm a_t$ for the original coefficients. With $K_{\rm eff}=\mu_1^2(C_n+b\Id_n)$, Eq.~\eqref{eq:appendix-flow-solution} becomes \begin{equation} 
\frac{d\widetilde{\bm a}_\tau}{d\tau} =y-(C_n+\beta\Id_n)\widetilde{\bm a}_\tau. 
\end{equation} 
For an eigenvector of $C_n$ with eigenvalue $s$, let $\widetilde a_j(\tau)$ and $y_j$ denote the projections of the coefficient vector and labels onto that eigenvector. Its kernel eigenvalue is $\mu_1^2(s+b)$, and its reduced dynamics are $d\widetilde a_j/d\tau=y_j-(s+\beta)\widetilde a_j$. Multiplying by the integrating factor $e^{\tau(s+\beta)}$ and integrating from $0$ to $\tau$ gives 
\begin{equation} 
\widetilde a_j(\tau)=\frac{1-e^{-\tau(s+\beta)}}{s+\beta}\,y_j +e^{-\tau(s+\beta)}\widetilde a_j(0). 
\label{eq:appendix-mode-solution} 
\end{equation} 
Define $r_\tau(s)=e^{-\tau(s+\beta)}$ and $A_\tau(s)=(1-r_\tau(s))/(s+\beta)$. Applying these scalar functions to each eigenvalue reconstructs the matrix solution: 
\begin{equation} 
\widetilde{\bm a}_\tau=A_\tau(C_n)y+r_\tau(C_n)\widetilde{\bm a}_0, \qquad \widetilde{\bm a}_0\sim\N(0,\delta_0\Id_n). 
\label{eq:scaled-solution} 
\end{equation}

These functions are usually denoted as spectral filters \citep{ali2019}: they specify how much each eigenvector contributes at time $\tau$. The training predictions are $(C_n+b\Id_n)\widetilde{\bm a}_\tau$, so their response to the labels is $h_\tau(s)=(s+b)A_\tau(s)$. For test prediction $f_t(x)=x^\top v_\tau/\sqrt d+o(1)$, the rule-dependent part of the labels contributes a factor $X/\sqrt d$ (through $g=Xw_\star/\sqrt d$), and forming the effective weight vector $v_\tau=X^\top\widetilde{\bm a}_\tau/\sqrt d$ in Eq.~\eqref{eq:test-transfer} contributes $X^\top/\sqrt d$. Together these give the factor $s$ in $F_\tau(s)=sA_\tau(s)$. 

\paragraph{The effective predictor and its overlap with the rule.} By Eq.~\eqref{eq:test-transfer}, $v_\tau=X^\top\widetilde{\bm a}_\tau/\sqrt d$, with the factor $\mu_1^2$ already included in $\widetilde{\bm a}_\tau$. Substituting the label decomposition in Eq.~\eqref{eq:signal-residual} into Eq.~\eqref{eq:scaled-solution} gives

\begin{equation} 
\begin{aligned} 
v_\tau &=a\frac{X^\top}{\sqrt d}A_\tau(C_n)\frac{X}{\sqrt d}w_\star +\frac{X^\top}{\sqrt d}A_\tau(C_n)\xi +\frac{X^\top}{\sqrt d}r_\tau(C_n)\widetilde{\bm a}_0\\ &=aF_\tau(C_d)w_\star +\frac{X^\top}{\sqrt d}A_\tau(C_n)\xi +\frac{X^\top}{\sqrt d}r_\tau(C_n)\widetilde{\bm a}_0, \qquad C_d=\frac{X^\top X}{d}. 
\label{eq:signal-noise-predictor} 
\end{aligned} 
\end{equation} 
To obtain the second line, write the compact singular-value decomposition $X/\sqrt d=US^{1/2}V^\top$, where $S$ contains the positive eigenvalues shared by $C_n$ and $C_d$. Then 
\begin{equation} 
\frac{X^\top}{\sqrt d}A_\tau(C_n)\frac{X}{\sqrt d} =VS^{1/2}A_\tau(S)S^{1/2}V^\top =VSA_\tau(S)V^\top =F_\tau(C_d). 
\end{equation} 
Each factor $X/\sqrt d$ or $X^\top/\sqrt d$ contributes $\sqrt s$, giving $sA_\tau(s)$. Both sides vanish on the null space of $X$ because $F_\tau(0)=0$. The three terms are the part fitted from the rule vector $w^\star$ , the part fitted from the label residual, and the remaining initialization.

The normalized overlap $v_\tau^\top w_\star/d$ measures alignment with the rule, and $\|v_\tau\|_2^2/d$ is the variance of predictions on a new input. They concentrate around the deterministic limits $m_\tau$ and $q_\tau$ at fixed $\alpha=n/d$ and $n,d\to\infty$. For example, multiplying Eq.~\eqref{eq:signal-noise-predictor} by $w_\star^\top/d$ and averaging gives 
\begin{equation} 
\frac1d\E[w_\star^\top v_\tau\mid X] =\frac{a}{d}\E[w_\star^\top F_\tau(C_d)w_\star\mid X] =\frac{a}{d}\Tr F_\tau(C_d). 
\label{eq:appendix-overlap-calculation} 
\end{equation} 
The residual term averages to zero by $\E[\xi w_\star^\top\mid X]=0$, and the initialization term averages to zero by independence. Similarly, the mean squared norm is the sum of the squared norms of the three terms in Eq.~\eqref{eq:signal-noise-predictor}: their pairwise scalar products have zero conditional mean. The three contributions are 
\begin{equation} 
\begin{aligned} 
\frac1d\E\|aF_\tau(C_d)w_\star\|_2^2 &=\frac{a^2}{d}\Tr F_\tau(C_d)^2,\\ \frac1d\E\left\|\frac{X^\top}{\sqrt d}A_\tau(C_n)\xi\right\|_2^2 &=\frac1d\E[\xi^\top A_\tau(C_n)C_nA_\tau(C_n)\xi] =\frac{1-a^2}{d}\Tr[C_nA_\tau(C_n)^2]+o(1),\\ \frac1d\E\left\|\frac{X^\top}{\sqrt d}r_\tau(C_n)\widetilde{\bm a}_0\right\|_2^2 &=\frac{\delta_0}{d}\Tr[C_nr_\tau(C_n)^2]. 
\label{eq:appendix-norm-calculation} 
\end{aligned} 
\end{equation} 
All expectations in these three lines are conditional on $X$. We used $\E[z^\top Mz]=\Tr[M\E(zz^\top)]$ and Eq.~\eqref{eq:label-covariance}. The $o(1)$ term comes from the covariance remainder in that equation. The normalized quadratic forms concentrate around these conditional means in the isotropic proportional limit \citep{advani2020,loureiro2021,bodin2022}. In particular, 
\begin{equation} 
\begin{aligned} 
m_\tau&=\lim_{n,d\to\infty}\frac{a}{d}\Tr F_\tau(C_d),\\ q_\tau&=\lim_{n,d\to\infty}\left\{ \frac{a^2}{d}\Tr F_\tau(C_d)^2 +\frac{1-a^2}{d}\Tr[C_nA_\tau(C_n)^2] +\frac{\delta_0}{d}\Tr[C_nr_\tau(C_n)^2]\right\}. 
\label{eq:overlap-traces} 
\end{aligned} 
\end{equation} 
These limits are taken at fixed $n/d\to\alpha$ as specified in Appendix~\ref{app:kernel} and Section~\ref{sec:setting} of the main text.

\paragraph{From matrix traces to one-dimensional integrals.} \label{app:spectra} The nonzero eigenvalues of $C_d$ and $C_n$ coincide. The eigenvalue distribution of $C_n$ tends to the Marchenko--Pastur law \citep{marchenko1967},

\begin{equation} 
\rho_\alpha(ds)=\left(1-\frac1\alpha\right)_+\delta_{\{0\}}(ds) +\frac{\sqrt{(s_+-s)(s-s_-)}}{2\pi\alpha s} \ind_{[s_-,s_+]}(s)\,ds, \qquad s_\pm=(1\pm\sqrt\alpha)^2. 
\label{eq:mp-measures} 
\end{equation} 
We use the same average notation as in the main text: $\langle\psi\rangle_\rho=\int\psi(s)\rho_\alpha(ds)$. Here $\rho_\alpha$ denotes the measure $\rho$ with its dependence on the sample ratio made explicit. It has total mass one and mean eigenvalue one, so $\langle1\rangle_\rho=\langle s\rangle_\rho=1$. Counting the same nonzero eigenvalues in the two matrices gives 
\begin{equation} 
\frac1d\Tr\psi(C_d)\longrightarrow\alpha\langle\psi\rangle_{\rho} \quad\text{when }\psi(0)=0, \qquad \frac1n\Tr\psi(C_n)\longrightarrow\langle\psi\rangle_{\rho}.
\label{eq:isotropic-measure-relation} 
\end{equation} 
For large $n,d$, the normalized trace $n^{-1}\Tr\psi(C_n)=n^{-1}\sum_{j=1}^n\psi(s_j)$ averages over the empirical eigenvalue distribution, which converges to $\rho_\alpha$. Changing the normalization from $n$ to $d$ introduces the factor $n/d\to\alpha$. Since $sA_\tau(s)^2=F_\tau(s)^2/s$, Eq.~\eqref{eq:overlap-traces} becomes 
\begin{equation} 
m_\tau=a\alpha\langle F_\tau\rangle_{\rho},\qquad q_\tau=\alpha\left\langle \left(a^2+\frac{1-a^2}{s}\right)F_\tau^2+\delta_0sr_\tau^2 \right\rangle_{\rho}. 
\label{eq:derived-mq} 
\end{equation} 
At $s=0$, $F_\tau^2/s$ is defined by its continuous limit, zero. These are Eq.~\eqref{eq:mq}. For a new spherical input, two linear classifiers disagree with probability equal to the angle between their weight vectors divided by $\pi$. Since $\|w_\star\|_2^2/d\to1$, the test error defined in the main text in Eq.~\eqref{eq:gen-def} therefore concentrates to the standard formula $\Egen(\tau)=\pi^{-1}\arccos(m_\tau/\sqrt{q_\tau})$, giving the first formula in Eq.~\eqref{eq:errors}.

\paragraph{Mean and variance of a factual training margin.} We now derive a low-dimensional expression for the memorization error on the training facts, defined in Eq.~\eqref{eq:mem-def}. We use the effective dynamics from Appendix~\ref{app:kernel}, which give the same limiting margin for a typical training sample. Their prediction vector is $f_t=(f_t(x^1),\ldots,f_t(x^n))^\top$ 
\begin{equation} 
f_t=K_{\rm eff}\bm a_t =(C_n+b\Id_n)\widetilde{\bm a}_\tau =h_\tau(C_n)y+(C_n+b\Id_n)r_\tau(C_n)\widetilde{\bm a}_0. 
\label{eq:appendix-training-predictions} 
\end{equation} 
For compactness within this calculation only, we write $B=h_\tau(C_n)$ and $J=(C_n+b\Id_n)r_\tau(C_n)$. The $\mu$th entry of Eq.~\eqref{eq:appendix-training-predictions} is $\sum_\nu B_{\mu\nu}y^\nu+(J\widetilde{\bm a}_0)_\mu$. For a fact $\mu\in\mathcal I_{\rm mem}$, multiply this entry by $y^\mu$ and separate the term $\nu=\mu$. Since $(y^\mu)^2=1$, the signed margin $y^\mu f_t(x^\mu)$ satisfies 
\begin{equation} 
y^\mu f_t(x^\mu)=B_{\mu\mu} +y^\mu\sum_{\nu\ne\mu}B_{\mu\nu}y^\nu +y^\mu(J\widetilde{\bm a}_0)_\mu+o(1). 
\label{eq:fact-self} 
\end{equation} 
Here $o(1)$ comes from replacing the full nonlinear-kernel predictor by its effective-kernel limit in Appendix~\ref{app:kernel}. We now compute the mean and variance of this margin. We hold the inputs $X$ fixed and condition on $\mu$ being a fact; the average is over the ground truth vector $w^\star$, the remaining label assignments, the random labels and the initialization. The order parameters in Eq.~\eqref{eq:VQ} are 
\begin{equation} 
\begin{aligned} 
V_\tau&=\lim_{n,d\to\infty} \E[y^\mu f_t(x^\mu)\mid X,\,\mu\in\mathcal I_{\rm mem}],\\ Q_\tau&=\lim_{n,d\to\infty} \Var(y^\mu f_t(x^\mu)\mid X,\,\mu\in\mathcal I_{\rm mem}).
\end{aligned} 
\label{eq:fact-margin-moments} 
\end{equation} 
The limits are in probability, with $n/d\to\alpha$ and fixed $\tau$. In Eq.~\eqref{eq:fact-self}, the fact's own label contributes $B_{\mu\mu}$, while the other two terms have zero mean: $y^\mu$ is a balanced sign independent of the remaining labels and initialization. Therefore \begin{equation} 
\E[y^\mu f_t(x^\mu)\mid X,\,\mu\in\mathcal I_{\rm mem}] =B_{\mu\mu}+o(1). 
\end{equation} 
For the variance, we subtract this mean and square the remaining terms. The cross term between the labels and initialization averages to zero because $\widetilde{\bm a}_0$ is centered and independent of the inputs, target vector and labels. Using $(y^\mu)^2=1$, we obtain 
\begin{equation} 
\begin{aligned} 
&\Var(y^\mu f_t(x^\mu)\mid X,\,\mu\in\mathcal I_{\rm mem})\\ 
&\quad=\E\!\left[(y^\mu f_t(x^\mu)-B_{\mu\mu})^2 \mid X,\,\mu\in\mathcal I_{\rm mem}\right]+o(1)\\ &\quad=\sum_{\nu,\kappa\ne\mu}B_{\mu\nu}B_{\mu\kappa}H_{\nu\kappa} +\delta_0(J^2)_{\mu\mu}+o(1)\\ 
&\quad=(1-a^2)\sum_{\nu\ne\mu}B_{\mu\nu}^2 +a^2\sum_{\nu,\kappa\ne\mu}B_{\mu\nu}C_{n,\nu\kappa}B_{\mu\kappa} +\delta_0(J^2)_{\mu\mu}+o(1). 
\end{aligned} 
\label{eq:fact-variance-matrix} 
\end{equation}

Here $H_{\nu\kappa}=\E[y^\nu y^\kappa\mid X]$ is the label covariance in Eq.~\eqref{eq:label-moments}, and $C_{n,\nu\kappa}=x^{\nu\top}x^\kappa/d$. Conditioning on $\mu$ being a fact leaves the other label moments unchanged, so Eq.~\eqref{eq:label-covariance} applies.

Removing the index $\mu$ from each sum gives the useful identities 
\begin{equation} 
\begin{aligned} 
\sum_{\nu\ne\mu}B_{\mu\nu}^2 &=(B^2)_{\mu\mu}-B_{\mu\mu}^2,\\ \sum_{\nu,\kappa\ne\mu}B_{\mu\nu}C_{n,\nu\kappa}B_{\mu\kappa} &=(BC_nB)_{\mu\mu} -2B_{\mu\mu}(BC_n)_{\mu\mu}+B_{\mu\mu}^2. 
\label{eq:fact-row-identities} 
\end{aligned} 
\end{equation} 
The last term uses $C_{n,\mu\mu}=1$.

For the independent spherical inputs, a typical diagonal entry of these matrix functions concentrates around its average, the normalized trace. With $B=h_\tau(C_n)=U\operatorname{diag}(h_\tau(s_j))U^\top$, $B_{\mu\mu}=\sum_{j=1}^nU_{\mu j}^2h_\tau(s_j)$. Isotropic eigenvectors have squared components averaging to $1/n$. The normalization is therefore $n$, the number of training examples, rather than $d$. Thus the mean gives \begin{equation} V_\tau =\lim_{\substack{n,d\to\infty\\ n/d\to\alpha}}B_{\mu\mu} =\lim_{\substack{n,d\to\infty\\ n/d\to\alpha}} \frac1n\Tr h_\tau(C_n) =\langle h_\tau\rangle_\rho. \label{eq:fact-mean-trace} \end{equation} The limits are in probability for a typical fixed index $\mu$ and fixed $\tau>0$. For the variance, the matrices $B^2$, $BC_n$, $BC_nB$ and $J^2$ have respective eigenvalues $h_\tau^2$, $s h_\tau$, $s h_\tau^2$ and $(s+b)^2r_\tau^2$. Replacing their diagonal entries in Eqs.~\eqref{eq:fact-variance-matrix}--\eqref{eq:fact-row-identities} by the corresponding spectral averages gives

\begin{equation} 
\begin{aligned} 
V_\tau&=\langle h_\tau\rangle_{\rho},\\ Q_\tau&=(1-a^2)\bigl(\langle h_\tau^2\rangle_{\rho}-V_\tau^2\bigr) +a^2\bigl(\langle sh_\tau^2\rangle_{\rho} -2V_\tau\langle sh_\tau\rangle_{{\rho}}+V_\tau^2\bigr) +\delta_0\langle(s+b)^2r_\tau^2\rangle_{\rho}\\ &=\left\langle(1-a^2+a^2s)(h_\tau-V_\tau)^2 +\delta_0(s+b)^2r_\tau^2\right\rangle_{\rho}. 
\end{aligned} 
\label{eq:fact-four-form}
\end{equation} 
The last equality uses $\langle s\rangle_{\rho}=1$ and gives Eq.~\eqref{eq:VQ}. The atom at zero in Eq.~\eqref{eq:mp-measures} must be included: when $\alpha>1$, the diagonal part of the kernel also fits training directions absent from the linear predictor on new inputs.

\paragraph{From factual margins to memorization error.} The moment calculation above does not by itself imply a Gaussian margin. We now justify this last step. For a typical fact $\mu$ and fixed $\tau>0$, the off-diagonal row of $B=h_\tau(C_n)$ is spread over many training examples. In particular, 
\begin{equation} 
\sum_{\nu\ne\mu}B_{\mu\nu}^2 \longrightarrow \langle h_\tau^2\rangle_\rho-V_\tau^2, \qquad \max_{\nu\ne\mu}|B_{\mu\nu}|\longrightarrow0. 
\label{eq:fact-row-spread} 
\end{equation} 
The first relation follows from Eq.~\eqref{eq:fact-row-identities} and the concentration of diagonal entries around normalized traces. For the second, recall the Gaussian representation of the spherical inputs from Appendix~\ref{app:kernel}, namely $x^\nu=\sqrt d\,z^\nu/\|z^\nu\|_2$ and $C_n=D(ZZ^\top/d)D$, with $D_{\nu\nu}=\sqrt d/\|z^\nu\|_2\to1$: the row normalization is asymptotically negligible, while the corresponding Gaussian sample covariance has isotropic eigenvectors because left multiplication of $Z$ by an orthogonal matrix leaves its Gaussian law unchanged. Thus the off-diagonal entries of the bounded spectral function $h_\tau(C_n)$ vanish individually. Equation \eqref{eq:fact-row-spread} says that the off-diagonal margin has finite total variance, but no individual training example contributes at order one.

To identify its limiting law, it is useful to separate the linear Gaussian part of the rule labels from the remaining fluctuations. Using $a=(1-\varepsilon)\sqrt{2/\pi}$, the off-diagonal label contribution in Eq.~\eqref{eq:fact-self} can be written exactly as 
\begin{equation} 
\begin{aligned} 
\sum_{\nu\ne\mu}B_{\mu\nu}y^\nu &=a\sum_{\nu\ne\mu}B_{\mu\nu}g^\nu\\ 
&\quad +(1-\varepsilon)\sum_{\nu\ne\mu}B_{\mu\nu} \left[\sign(g^\nu)-\sqrt{\frac{2}{\pi}}\,g^\nu\right]\\ 
&\quad +\sum_{\nu\ne\mu}B_{\mu\nu} \left[y^\nu-(1-\varepsilon)\sign(g^\nu)\right]. 
\end{aligned} 
\label{eq:fact-gaussian-decomposition} 
\end{equation} 
We added and subtracted $(1-\varepsilon)\sign(g^\nu)$, then removed its linear Gaussian projection $ag^\nu$. This projection follows from $\E[G\sign G]=\sqrt{2/\pi}$, so $\E[G(\sign G-\sqrt{2/\pi}\,G)]=0$. The first term is a linear function of the Gaussian field $g$, with $g^\nu=x^{\nu\top}w_\star/\sqrt d$, and is therefore Gaussian exactly.

The last term is the fluctuation of each label around its mean when the rule field is known. Conditional on $g$, the label agrees with $\sign(g^\nu)$ with probability $1-\varepsilon/2$ and has the opposite sign with probability $\varepsilon/2$. Hence
\begin{equation}
\E[y^\nu\mid g]=(1-\varepsilon)\sign(g^\nu),\qquad \Var(y^\nu\mid g)=\varepsilon(2-\varepsilon).
\end{equation}
These centered fluctuations are independent across examples at fixed $g$ and are bounded by two in absolute value. Their weighted sum has variance $\varepsilon(2-\varepsilon)\sum_{\nu\ne\mu}B_{\mu\nu}^2$. Equation~\eqref{eq:fact-row-spread} makes this variance finite while the largest individual contribution vanishes, giving a Gaussian limit by the conditional central-limit theorem. Its limiting variance does not depend on $g$, so this contribution is asymptotically independent of the first two terms.

The middle term in Eq.~\eqref{eq:fact-gaussian-decomposition} contains the rule labels after their linear Gaussian part has been subtracted. For a standard Gaussian $G$, the residual $\sign(G)-\sqrt{2/\pi}\,G$ has zero correlation with $G$ and variance $1-2/\pi$. The bounded spectrum of $C_n$ and the weight estimates in Eq.~\eqref{eq:fact-row-spread} allow the Gaussian central-limit criteria to be applied to polynomial approximations of this residual, following \citet{nualartpeccati2005,peccatitudor2005}. 
As the polynomial approximation is refined, the mean squared difference between the weighted residual in the second line of Eq.~\eqref{eq:fact-gaussian-decomposition} and its polynomial approximation tends to zero. The weighted residual therefore has a joint Gaussian limit with the linear Gaussian term in the first line of the same equation. Their covariance is zero because $\sign(g^\nu)-\sqrt{2/\pi}\,g^\nu$, the residual appearing in the second line, is uncorrelated with every component of $g$. The two limiting Gaussian variables are therefore independent.
Their variances, together with that of the last term, give the label contribution already calculated in Eq.~\eqref{eq:fact-four-form}. Finally, the initialization contribution $y^\mu(J\widetilde{\bm a}_0)_\mu$ in Eq.~\eqref{eq:fact-self} is already Gaussian conditional on $X$, since $\widetilde{\bm a}_0$ is Gaussian and independent of the data and labels. For a fact, $y^\mu$ is an independent balanced sign, so this multiplication does not change the centered Gaussian law.

Combining these contributions, the signed margin of a typical fact converges in distribution, at fixed positive $\tau$, as $n,d\to\infty$ with $n/d\to\alpha$: 
\begin{equation} 
y^\mu f_t(x^\mu) \longrightarrow V_\tau+\sqrt{Q_\tau}\,Z, \qquad Z\sim\N(0,1), \qquad t=\tau/\mu_1^2. 
\label{eq:fact-margin-limit} 
\end{equation}
Here $V_\tau$ and $Q_\tau$ are the limiting mean and variance derived in Eq.~\eqref{eq:fact-four-form}.

It remains to pass from a typical fact to the empirical memorization error. Let $I_\mu=\ind\{y^\mu f_t(x^\mu)<0\}$ for a factual example. The measured error is the average $\Emem=|\mathcal I_{\rm mem}|^{-1}\sum_{\mu\in\mathcal I_{\rm mem}}I_\mu$. For two distinct facts, the joint Gaussian margin fluctuations have zero limiting covariance: the cross terms are off-diagonal entries of the bounded matrix functions used above, and these vanish. Their misclassification indicators therefore become uncorrelated. The variance of their average consists of a diagonal contribution of order $1/|\mathcal I_{\rm mem}|$ and a vanishing off-diagonal contribution. Since $|\mathcal I_{\rm mem}|/n\to\varepsilon>0$, the measured fraction converges to the single-fact Gaussian tail probability, 
\begin{equation}
\Emem(\tau) =\Prob\!\left(V_\tau+\sqrt{Q_\tau}\,Z<0\right) =\frac12\erfc\!\left( \frac{V_\tau}{\sqrt{2Q_\tau}} \right). 
\label{eq:derived-mem-error} 
\end{equation}

\paragraph{Final low-dimensional equations.} Combining Eqs.~\eqref{eq:derived-mq}, \eqref{eq:fact-four-form} and~\eqref{eq:derived-mem-error} recovers the main formulas:
\begin{equation} 
\begin{aligned} 
m_\tau&=a\alpha\langle F_\tau\rangle_{\rho}, &q_\tau&=\alpha\left\langle \left(a^2+\frac{1-a^2}{s}\right)F_\tau^2+\delta_0sr_\tau^2\right\rangle_{\rho},\\ V_\tau&=\langle h_\tau\rangle_{\rho}, &Q_\tau&=\left\langle(1-a^2+a^2s)(h_\tau-V_\tau)^2 +\delta_0(s+b)^2r_\tau^2\right\rangle_{\rho},\\ \Egen(\tau)&=\frac1\pi\arccos\!\left(\frac{m_\tau}{\sqrt{q_\tau}}\right), &\Emem(\tau)&=\frac12\erfc\!\left(\frac{V_\tau}{\sqrt{2Q_\tau}}\right). 
\end{aligned} 
\label{eq:appendix-final-curves}
\end{equation} 
Here again $a=(1-\varepsilon)\sqrt{2/\pi}$, $n/d\to\alpha$, $\delta_0 = \mu_1^4\Delta_0$ initialization variance. All averages are one-dimensional Marchenko--Pastur integrals. The filters are given by Eq.~\eqref{eq:filters}, with $b=\cot^2\gamma$ and the reduced quantities in Eq.~\eqref{eq:appendix-reduced}. These equations are Eqs.~\eqref{eq:errors}, \eqref{eq:mq} and~\eqref{eq:VQ} of the main text. At zero ridge and convergence, $h_\infty=1$ and $r_\infty=0$, so $V_\infty=1$ and $Q_\infty=0$. Every factual margin then tends to one and $\Emem(\infty)=0$, consistently with zero square loss on the whole training set. Generalization still depends on the kernel angle through $F_\infty(s)=s/(s+\cot^2\gamma)$.

\section{Why the critical kernel separates the stopping regimes} \label{app:proof}\label{app:boundary} 

In this Appendix section we use $c(\varepsilon)=a/\sqrt{1-a^2}$, $\bc=(1-a^2)/a^2=c(\varepsilon)^{-2}$, and $\gamma_{\rm c}(\varepsilon)=\arctan c(\varepsilon)$. We first prove the stopping criterion of Section~\ref{sec:errors}, Eq.~\eqref{eq:boundary}. At zero ridge, the test error initially decreases for every $\gamma$; it keeps decreasing when $\gamma\le\gamma_{\rm c}(\varepsilon)$ and eventually increases when $\gamma>\gamma_{\rm c}(\varepsilon)$. We then compare all kernel angles and training times, proving the bound in Section~\ref{sec:stopping}, Eq.~\eqref{eq:global-optimum}, and showing that the critical kernel attains it at interpolation. Finally, we check that Euclidean coefficient flow has the same stopping boundary. We consider $0\le\varepsilon<1$, $b>0$, and fixed finite initialization variance $\delta_0\ge0$. All statements concern the limiting curves obtained as $n,d\to\infty$ at fixed $n/d\to\alpha>0$; time is varied after this limit. Write $\beta=b+\ell$. Separating the label and initialization contributions to $q=q_L+q_I$, and writing $M=m/a$, gives

\begin{equation}
\begin{aligned} 
M&=\alpha\left\langle\frac{s}{s+\beta}(1-r)\right\rangle_{\rho}, &q_L&=a^2\alpha\left\langle\frac{s(s+\bc)}{(s+\beta)^2}(1-r)^2\right\rangle_{\rho},\\ 
q_I&=\delta_0\alpha\langle sr^2\rangle_{\rho}, &r&=e^{-\tau(s+\beta)},\\ 
q&=q_L+q_I, &M&=m/a. 
\end{aligned} 
\label{eq:proof-defs} 
\end{equation}

Let $\tau_{\rm es}$ minimize $\Egen(\tau)$ over $\tau\in(0,\infty]$, with $t_{\rm es}=\tau_{\rm es}/\mu_1^2$. Since $\arccos$ decreases and $m_\tau>0$ for $\tau>0$, this is equivalent to maximizing $m_\tau/\sqrt{q_\tau}$, or its square $\chi=m^2/q$. Differentiating Eq.~\eqref{eq:appendix-final-curves} gives 
\begin{equation} 
\Egen'(\tau)=-\frac{2m' q-mq'}{2\pi q^{3/2}\sqrt{1-m^2/q}}. 
\end{equation} 
Thus every minimum at a finite positive time satisfies $2m'q-mq'=0$. We compare all candidate minima with the endpoints; uniqueness is not assumed.

\paragraph{Test error initially decreases for every kernel angle.} This step rules out the initial classifier as the best point along the trajectory. If $\delta_0>0$, Eq.~\eqref{eq:proof-defs} gives $m=a\alpha\tau+O(\tau^2)$ and $q=\delta_0\alpha+O(\tau)$, so $\chi\sim a^2\alpha\tau^2/\delta_0$ increases just after initialization. If $\delta_0=0$, both $m$ and $q$ vanish at $\tau=0$, and we must start from the one-sided classification limit $\tau\to0^+$. Expanding $1-r=\tau(s+\beta)-\tfrac12\tau^2(s+\beta)^2+O(\tau^3)$ in Eq.~\eqref{eq:proof-defs} gives 
\begin{equation} 
\begin{aligned} 
\left.\partial_\tau\log\chi\right|_{0^+} &=\frac{\langle s(s+\bc)(s+\beta)\rangle_\rho} {\langle s(s+\bc)\rangle_\rho} -\frac{\langle s(s+\beta)\rangle_\rho}{\langle s\rangle_\rho}\\ 
&=\frac{\alpha}{1+\alpha+\bc}>0. 
\end{aligned}
\label{eq:initial-improvement} 
\end{equation} 
The last equality uses the Marchenko--Pastur moments $\langle s\rangle_\rho=1$, $\langle s^2\rangle_\rho=1+\alpha$ and $\langle s^3\rangle_\rho=1+3\alpha+\alpha^2$. Thus $\Egen$ initially decreases for either initialization and for every $\gamma$. We now determine whether its derivative remains negative or eventually becomes positive.

\paragraph{Convergence is optimal for $\gamma\le\gamma_{\rm c}(\varepsilon)$ at zero ridge.} At $\lambda=0$, $\gamma\le\gamma_{\rm c}(\varepsilon)$ is equivalent to $b\ge\bc$. The same proof also covers positive ridge whenever $\beta=\cot^2\gamma+\ell\ge\cot^2\gamma_{\rm c}(\varepsilon)=\bc$. Use the existing filter $F_\tau(s)=s(1-r)/(s+\beta)$ defined in Appendix~\ref{app:derivations}. In Eq.~\eqref{eq:proof-defs}, an eigenvalue contributes $F_\tau(s)$ to $M/\alpha$ and $F_\tau(s)R_\tau(s)$ to $q_L/\alpha$, where 
\begin{equation} 
R_\tau(s)=a^2\frac{s+\bc}{s+\beta}(1-r),\qquad L_\tau(s)=\frac{\partial_\tau F_\tau(s)}{F_\tau(s)} =\frac{s+\beta}{e^{\tau(s+\beta)}-1}. 
\label{eq:boundary-weights} 
\end{equation} 
Thus $R_\tau$ is the ratio of the integrands of $q_L/\alpha$ (the label-induced prediction variance) and $M/\alpha$ (the rescaled rule alignment), and $L_\tau$ is the fractional rate at which the latter integrand grows. For $\beta\ge\bc$, $R_\tau$ increases with $s$, whereas $L_\tau$ decreases.

For completeness, normalize the signal contributions into weights by defining $\E_F[g]=\langle F_\tau g\rangle_{\rho}/\langle F_\tau\rangle_{\rho}$. Since $\partial_\tau F_\tau=F_\tau L_\tau$ and $\partial_\tau R_\tau=R_\tau L_\tau$, differentiation gives 
\begin{equation} 
M'=\alpha\langle F_\tau L_\tau\rangle_\rho,\qquad q_L'=2\alpha\langle F_\tau R_\tau L_\tau\rangle_\rho. 
\label{eq:boundary-rate-identities} 
\end{equation} 
Then $M'/M=\E_F[L_\tau]$ and $q_L'/(2q_L)=\E_F[R_\tau L_\tau]/\E_F[R_\tau]$, so 
\begin{equation} 
\frac{q_L'}{2q_L}-\frac{M'}M =\frac{\operatorname{Cov}_F(R_\tau,L_\tau)}{\E_F[R_\tau]}\le0. 
\label{eq:covariance-proof}
\end{equation}
Subtracting the two rates gives the numerator $\operatorname{Cov}_F(R_\tau,L_\tau)=\E_F[R_\tau L_\tau]-\E_F[R_\tau]\E_F[L_\tau]$. Here $\operatorname{Cov}_F$ is covariance under those same weights. Its sign follows directly from $(R_\tau(s)-R_\tau(s'))(L_\tau(s)-L_\tau(s'))\le0$ for every pair of eigenvalues, since $R_\tau$ increases while $L_\tau$ decreases. Therefore $m^2/q_L$ increases, since $\partial_\tau\log(m^2/q_L)=2M'/M-q_L'/q_L\ge0$. Initialization adds a variance $q_I$ that decreases with time, while $q_L$ increases: 
\begin{equation} 
\partial_\tau\log\frac{M^2}{q_L+q_I} -\partial_\tau\log\frac{M^2}{q_L} =\frac{q_L'q_I-q_Lq_I'}{q_L(q_L+q_I)}\ge0. 
\label{eq:init-monotonic-boundary} 
\end{equation} 
Its decay therefore reinforces the improvement in generalization. The Marchenko--Pastur law has an interval of positive eigenvalues, so the covariance is strictly negative at every finite positive time: unequal positive eigenvalues give a strictly negative product above, and their spectral weights are positive. Hence the test error strictly decreases until convergence, including at $\beta=\bc$. At zero ridge, this is exactly the regime $\gamma\le\gamma_{\rm c}(\varepsilon)$.

\paragraph{Finite stopping is beneficial for $\gamma>\gamma_{\rm c}(\varepsilon)$ at zero ridge.} At zero ridge, this is the regime $b=\cot^2\gamma<\cot^2\gamma_{\rm c}(\varepsilon)=\bc$. More generally, suppose $\beta=\cot^2\gamma+\ell<\bc$. The long-time variance-to-signal ratio $a^2(s+\bc)/(s+\beta)$ decreases with $s$. The smallest positive eigenvalues of $C_n$ (corresponding to eigenvalues $\mu_1^2(s+b)$ of $K_{\rm eff}$) therefore carry the most variance per unit signal, and they are also the last to relax. Let $s_0$ be the lower edge of the positive spectrum, possibly zero at $\alpha=1$. At long times, using $\partial_\tau F=sr$ gives the weighted average below. Its weight $sr=s e^{-\tau(s+\beta)}$ concentrates near $s_0$ as $\tau\to\infty$, since larger eigenvalues have an additional exponential decay.
\begin{equation} 
\frac{q_L'}{2M'}=\frac{\langle srR_\tau\rangle_\rho}{\langle sr\rangle_\rho}\longrightarrow a^2\frac{s_0+\bc}{s_0+\beta} >\frac{q_{L,\infty}}{M_\infty}. 
\label{eq:late-boundary} 
\end{equation} 
By contrast, $q_{L,\infty}/M_\infty=\langle F_\infty R_\infty\rangle_\rho/\langle F_\infty\rangle_\rho$ averages $R_\infty(s)=a^2(s+\bc)/(s+\beta)$ over the full positive spectrum. Since this function strictly decreases, its average is smaller than its value at $s_0$. Consequently $\chi$ eventually decreases, so the test error $\Egen$ eventually increases. The initialization has already decayed too quickly to change this conclusion: $|q_I'|/M'$ is bounded by a constant times $e^{-\tau\beta}$.

Together with the initial decrease proved above, this late increase places the minimum test error at a finite positive time. This establishes the criterion in Section~\ref{sec:errors}, Eq.~\eqref{eq:boundary}: $\beta<\bc$ is equivalent to $\mu_1>c(\varepsilon)\sqrt{\mu_\star^2+\lambda}$, or $\gamma>\gamma_{\rm c}(\varepsilon)$ at zero ridge. The argument does not require a unique turning point.

\paragraph{Finding the finite stopping time.} \label{app:scalar} The derivatives follow directly from Eq.~\eqref{eq:derived-mq}. Since $\partial_\tau r=-(s+\beta)r$ and $\partial_\tau F=sr$, differentiation under the spectral integral gives 
\begin{equation} 
m'=a\alpha\langle\partial_\tau F\rangle_\rho,\qquad q'=2\alpha\left\langle \left(a^2+\frac{1-a^2}{s}\right)F\partial_\tau F +\delta_0sr\partial_\tau r\right\rangle_\rho. 
\end{equation} 
Using $F=s(1-r)/(s+\beta)$ then gives the explicit expressions 
\begin{equation} 
\begin{aligned} 
m'&=a\alpha\langle sr\rangle_{\rho},\\ q'&=2\alpha\left\langle \frac{s(a^2s+1-a^2)}{s+\beta}(1-r)r -\delta_0s(s+\beta)r^2\right\rangle_{\rho}. 
\end{aligned} 
\label{eq:stopping-derivatives} 
\end{equation} 
We compare the errors at the stationary candidates and the endpoints. A finite observation window cannot by itself establish that the optimum is at infinity; the boundary above does.

\paragraph{The critical kernel attains the lowest test error across angles and times.} \label{app:equilibrium} The stopping criterion compares times for a fixed kernel. We now set $\lambda=0$ and compare the best test errors across all kernel geometries, parametrized by the single angle $\gamma$, as in Section~\ref{sec:stopping}. Starting from Eq.~\eqref{eq:derived-mq} or the trace formulas in Eq.~\eqref{eq:overlap-traces}, drop the nonnegative initialization variance and apply Cauchy--Schwarz. Writing $W(s)=1+\bc/s$, the two steps are 
\begin{equation} 
\begin{aligned} 
q&=a^2\alpha\langle WF^2\rangle_\rho+q_I\ge a^2\alpha\langle WF^2\rangle_\rho,\\ \langle F\rangle_\rho^2&=\left\langle(\sqrt W F)\frac1{\sqrt W}\right\rangle_\rho^2\le\langle WF^2\rangle_\rho\left\langle\frac1W\right\rangle_\rho. 
\end{aligned} 
\label{eq:bound-two-steps} 
\end{equation} 
Combining them gives 
\begin{equation} 
\frac{m^2}{q} \le\frac{\alpha\langle F\rangle_{\rho}^2} {\langle(1+\bc/s)F^2\rangle_{\rho}} \le I_\star:=\alpha\left\langle\frac{s}{s+\bc}\right\rangle_{\rho}, \qquad \Egen\ge\frac1\pi\arccos\sqrt{I_\star}. 
\label{eq:spectral-optimum} 
\end{equation} 
This is the bound in Eq.~\eqref{eq:global-optimum}; it holds for every kernel angle and every training time. Here $I_\star$ denotes the resulting upper bound on squared correlation, and $\arccos$ is decreasing. Equality requires zero surviving initialization variance and $F(s)$ proportional to $s/(s+\bc)$, as stated in Eq.~\eqref{eq:main-equality-condition}. At convergence, $F_\infty(s)=s/(s+\cot^2\gamma)$, so this condition holds only at $\gamma=\gamma_{\rm c}(\varepsilon)$. There $V_\infty=1$ and $Q_\infty=0$, and the predictor interpolates. Thus the critical endpoint attains the lowest test error across all angles and all training times, extending the ERM comparison of \citet{farne2026} to the full learning trajectory. With the argument below we now show that no finite stopping time attains the same value.

At finite $\tau>0$, equality would require $(s+\bc)(1-e^{-\tau(s+b)})=C(s+b)$ throughout the positive spectral interval. The right-hand side is linear in $s$, whereas the second derivative of the left-hand side is $\tau e^{-\tau(s+b)}[2-\tau(s+\bc)]$, which cannot vanish on an interval. Thus no finite-time filter attains equality, even at zero initialization. In particular,
\begin{equation} 
\Egen(\taue(\gamma);\gamma)>\Egen(\infty;\gamma_{\rm c}(\varepsilon))\quad\text{for every }\gamma>\gamma_{\rm c}(\varepsilon).
\label{eq:strict-stopping-gap}
\end{equation}
Positive ridge can instead achieve $\beta=\bc$ at convergence when $b<\bc$ \citep{farne2026}; this matches the optimal test error but no longer interpolates the training scores. All optimality statements here concern the stated smooth dot-product kernel family and training trajectories.

\paragraph{RKHS and Euclidean coefficient flow have the same boundary.} \label{app:metric} The two flows minimize the same square-loss objective and converge to the same coefficients $(K+\lambda\Id_n)^{-1}y$. Their distinction is the metric used for the gradient. In coefficient coordinates, RKHS length is $d\bm a^\top K\,d\bm a$, so the RKHS gradient multiplies the ordinary coefficient derivative by $K^{-1}$: 
\begin{equation}
\nabla_{\bm a}L_\lambda =K[(K+\lambda\Id_n)\bm a-y],\qquad -K^{-1}\nabla_{\bm a}L_\lambda =y-(K+\lambda\Id_n)\bm a. 
\label{eq:coefficient-metric} 
\end{equation} 
Consequently, Euclidean coefficient flow has one additional factor $K$ in its dynamics. In the spectral formulas, the only change is the exponential factor and the time scale: 
\begin{equation} 
\begin{aligned} 
\text{RKHS flow: }&r_\tau(s)=e^{-\tau(s+\beta)}, &&\tau=\mu_1^2t,\\ \text{Euclidean coefficient flow: }&r_\tau(s)=e^{-\tau(s+b)(s+\beta)}, &&\tau=\mu_1^4t. 
\end{aligned} 
\label{eq:metric-replacement} 
\end{equation} 
Substituting the second exponential into Eq.~\eqref{eq:appendix-final-curves} gives the Euclidean learning curves. The curves and optimal times generally differ, but their ERM errors coincide because both exponentials vanish at convergence.

The stopping proof also keeps the same sign. For $\beta\ge\bc$, $R_\tau$ in Eq.~\eqref{eq:boundary-weights} is still increasing, and the fractional rate is now $z/(e^{\tau z}-1)$ with $z=(s+b)(s+\beta)$, which still decreases with $s$. For $\beta<\bc$, the smallest positive eigenvalues still relax last and have the largest limiting variance per unit signal. The initialization contribution decays faster than this late-time correction. Initial improvement at $\delta_0=0$ follows by the same expansion, with an increasing variance-to-signal weight proportional to $(s+\bc)(s+b)$; at $\delta_0>0$, $m^2/q$ again starts proportionally to $\tau^2$. Thus both metrics have the boundary $b+\ell=\bc$, or $\gamma=\gamma_{\rm c}(\varepsilon)$ at zero ridge. The main figures use RKHS flow.

\section{Extension to a general input covariance} \label{app:covariance} In this appendix we show why the boundary survives an anisotropic input distribution when the target vector remains independent and isotropic. Replace only the input model by 
\begin{equation} 
x^\mu=\frac{\sqrt d\,\Sigma^{1/2}z^\mu}{\|\Sigma^{1/2}z^\mu\|_2},\qquad z^\mu\sim\N(0,\Id_d),\qquad \Tr\Sigma=d,\qquad \sup_d\|\Sigma\|_{\op}<\infty, 
\label{eq:design-assumptions} 
\end{equation} 
and suppose the eigenvalue distribution of $\Sigma$ tends to a probability law $H$. The target vector is still $w_\star\sim\N(0,\Id_d)$. The normalization fixes the rule field variance to one and gives inputs with asymptotic covariance $\Sigma$; their finite-dimensional covariance need not be exactly $\Sigma$. Bounded eigenvalues exclude a single input direction carrying a finite fraction of the total variance. Here the rule field is $g_\star(x)=x^\top w_\star/\sqrt d$, and its variance over the independent target vector at fixed input is $\E_{w_\star}[g_\star(x)^2\mid x]=\|x\|_2^2/d=1$. The trace condition fixes only the scale: multiplying $\Sigma$ by a positive constant leaves the normalized inputs unchanged.

The kernel expansion \eqref{eq:full-kernel-reduction} remains valid with quadratic constant $\sigma''(0)\Tr\Sigma^2/(2d^2)$ \citep[Theorem~2.1]{elkaroui2010}. Gaussian sample-space rotational invariance, followed by asymptotically negligible row normalization, gives the same diagonal and factual-margin calculations. Hence Eq.~\eqref{eq:VQ} remains valid with $\rho$ equal to the new eigenvalue law of $C_n$.

Test errors require an additional weighting: a new input has different variance in different feature directions. Define the positive measure $\nu$ by 
\begin{equation} 
\int\psi(s)\nu(ds)=\lim_{\substack{n,d\to\infty\\n/d\to\alpha}} \frac1d\Tr[\Sigma\psi(C_d)],\qquad C_d=X^\top X/d. 
\label{eq:nu}
\end{equation} 
It has total mass one. Repeating the trace calculation with test overlaps $m=d^{-1}v^\top\Sigma w_\star$ and $q=d^{-1}v^\top\Sigma v$ yields 
\begin{equation} 
\begin{aligned} 
m_\tau&=a\int F_\tau(s)\nu(ds),\\ q_\tau&=\int\left[\left(a^2+\frac{1-a^2}{s}\right)F_\tau(s)^2 +\delta_0sr_\tau(s)^2\right]\nu(ds). 
\end{aligned} 
\label{eq:general-covariance-overlaps}
\end{equation} 
Cyclicity of the trace suffices; $\Sigma$ need not commute with $C_d$. For isotropic inputs, Eq.~\eqref{eq:isotropic-measure-relation} recovers the main formulas. In general $\nu$ differs from both the ordinary feature eigenvalue law and $\alpha\rho$.

For numerical evaluation, one scalar covariance resolvent equation is sufficient \citep{silverstein1995,hachem2013}. In particular let $m_\rho(z)=\int(s-z)^{-1}\rho(ds)$ be the Stieltjes transform of the eigenvalue law of $C_n$. With the normalization $C_n=XX^\top/d$, it satisfies
\begin{equation}
z=-\frac1{m_\rho(z)}+\int\frac{u}{1+\alpha u m_\rho(z)}H(du),\qquad \operatorname{Im}z>0.
\label{eq:silverstein}
\end{equation}
For $\operatorname{Im}z>0$, choose the solution with positive imaginary part and $m_\rho(z)\sim-1/z$ at infinity. The continuous eigenvalue density is
\begin{equation}
\rho_{\rm cont}(s)=\frac1\pi\lim_{\eta\downarrow0}\operatorname{Im}m_\rho(s+i\eta).
\end{equation}
Any zero-eigenvalue mass is included separately in the averages defining $V_\tau,Q_\tau$. The test measure in Eq.~\eqref{eq:nu} follows from the same solution: on the positive spectrum,
\begin{equation}
\nu(ds)=\frac{\rho(ds)}{s|m_\rho(s+i0)|^2}.
\end{equation}
Thus only one density is computed, with an additional weight in the test integrals of Eq.~\eqref{eq:general-covariance-overlaps}. This weight accounts for the variance of a new input along each sample eigenvector. A possible atom of $\nu$ at zero does not contribute there, since the test integrands vanish at zero. For isotropic inputs the weight is $\alpha$, recovering the main formulas.

\paragraph{The same boundary, different errors and times.} Replace $\alpha\langle\cdot\rangle_{\rho}$ by $\int\cdot\,d\nu$ throughout the stopping proof. For initial improvement, use the spectral-moment expression in Eq.~\eqref{eq:initial-improvement}, before its Marchenko--Pastur simplification. Positivity of the measure, the opposing order of $R$ and $L$, and the late-time concentration on the smallest supported positive eigenvalues are unchanged. Hence, for a nondegenerate positive spectrum, the boundary remains $b+\ell=\bc$, and the best squared correlation becomes $I_\star=\int s/(s+\bc)\,\nu(ds)$, again attained by the unregularized critical kernel. An interval of positive spectral support is sufficient to exclude finite-time equality; a discrete spectrum requires checking the filter equality at its supported eigenvalues. A single positive eigenvalue only rescales a zero-initialized classifier. These conclusions retain $b>0$ and finite initialization variance. They do not cover a target vector aligned with selected covariance eigenvectors, covariance spikes growing with dimension, or arbitrary structured inputs: those changes can give different spectral weights to rule learning and fitting random labels and can alter the boundary.

\section{Numerical checks and kernel experiments} \label{app:numerical-checks}

In this section, we test our theoretical predictions in the previous appendix sections and in the main text against numerical simulations in several different learning regimes. We first compare the predicted generalization and memorization curves with the dynamics of a full nonlinear RBF kernel on synthetic RAF data. We then move to the CIFAR10 dataset construction of the main text, namely with cat/dogs images preserving their true labels (to generalize on) and randomized automobiles (to memorize), and examine whether changing the kernel produces qualitatively similar behavior on real image data.

\subsection{Numerical evaluation of the gradient-flow and ridge comparison} \label{app:time-ridge-numerics} Figure~\ref{fig:gf-ridge-difference} compares two numerical evaluations of the analytical learning curves for kernel ridge regression and for kernel gradient flow. We use Eqs.~\eqref{eq:errors}--\eqref{eq:VQ} at $\lambda=0$ for gradient flow and their $\tau\to\infty$ limit at positive $\lambda$ for ridge, consistent with the ERM analysis of \citet[Eqs.~(21), (24), (35) and (36)]{farne2026}.

\subsection{Validation with the full RBF kernel} \label{app:theory-numerics} We generate data from the spherical RAF model and use the kernel 
\begin{equation} 
k_\eta(x,x')=\exp\!\left[-\eta\left(1-\frac{x^\top x'}d\right)\right], \qquad \|x\|_2^2=\|x'\|_2^2=d. 
\label{eq:validation-rbf} 
\end{equation} 
For each theoretical kernel angle, we choose the bandwidth $\eta$ through Eq.~\eqref{eq:rbf-masses}. The simulations use the full training and test kernels, retaining all nonlinear contributions and the nonzero constant term $\mu_0^2=e^{-\eta}$. No centering or subtraction of this term is performed. This allows us to test the effective-kernel reduction as well as the learning curves derived from it.

\begin{figure}[!htbp] \centering \includegraphics[width=0.88\linewidth]{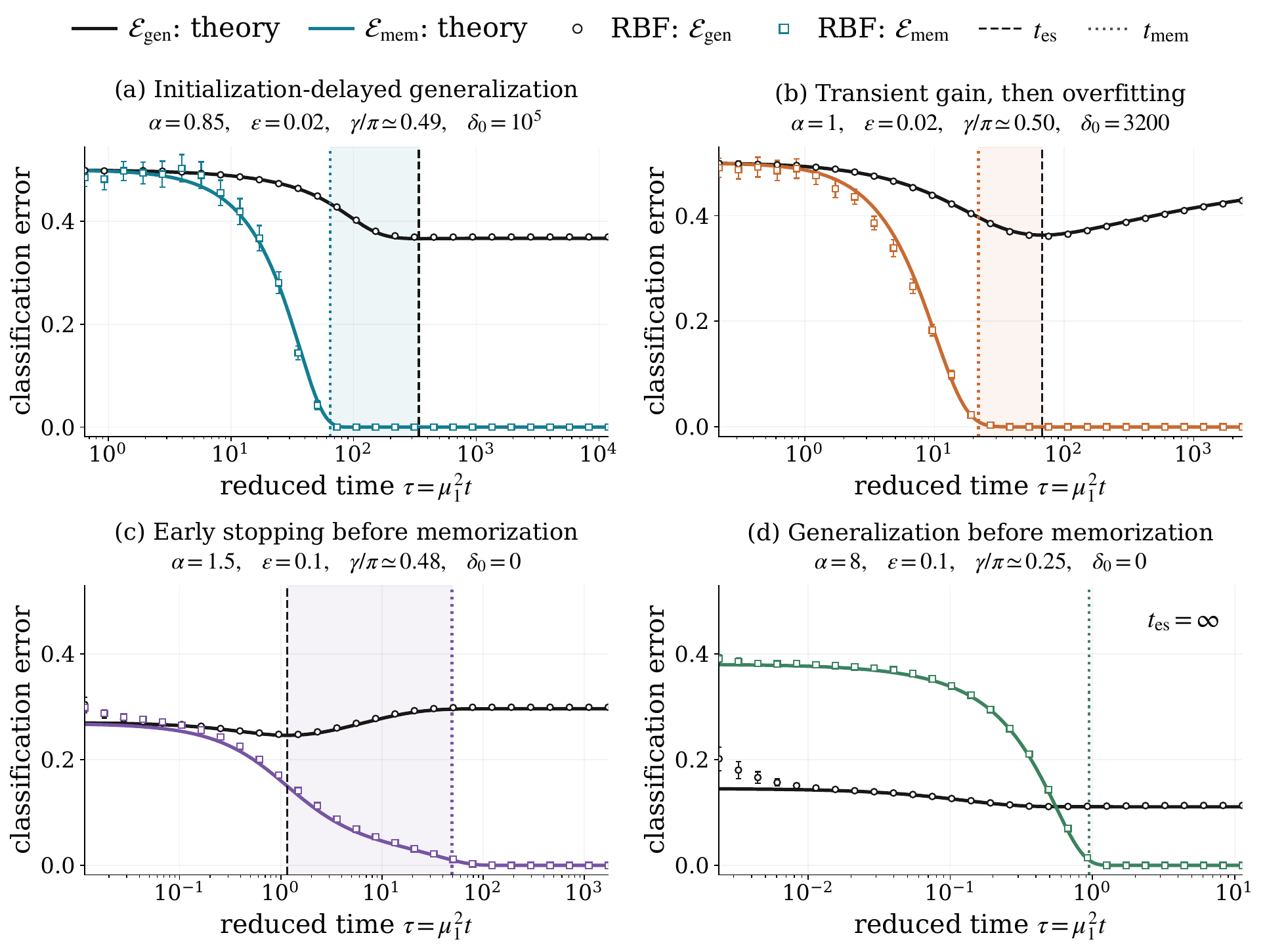} \caption{\textbf{Numerical validation across four learning regimes.} Lines show the theoretical generalization and memorization errors; circles and squares show the corresponding means from RBF simulations, with one standard error bars. All panels use $\lambda=0$ and retain the kernel's nonzero constant term. Parameters are shown above the panels. (a) Delayed generalization after factual memorization. (b) Delayed generalization followed by late overfitting. (c) Optimal stopping before factual memorization. (d) Continued improvement toward the critical interpolating endpoint. Dotted lines mark $\tmem(0.01)$ and dashed lines finite $\tes$. The corresponding reduced times are $\taue=\mu_1^2\tes$ and $\taum(\delta)=\mu_1^2\tmem(\delta)$.} \label{fig:regimes} \end{figure}

We compute the finite-sample dynamics by diagonalizing the RBF matrix and evaluating the high dimensional predictor in Eq.~\eqref{eq:finite-flow}, with $\lambda=0$. Initialization is matched to the theory through $\mu_1^2\bm a_0\sim\N(0,\delta_0\Id_n)$. The theoretical curves are evaluated by spectral quadrature, including the zero-eigenvalue mass, and checked at higher resolution. Figure~\ref{fig:regimes} uses $d=4000$ in panels (a,b,c) and $d=1000$ in panel (d). Each panel averages 12 independent realizations of the data, ground-truth vector $w_\star$ and initialization, with 10000 fresh test inputs per realization. The remaining parameters are indicated in the figure.

Across these choices, the simulations reproduce the predicted learning curves, including delayed improvements in generalization and late overfitting. The same experiments also illustrate how the parameters change the relative timing of rule learning and factual memorization. Alongside the optimal stopping time $\tes$ in Eq.~\eqref{eq:optimal_stopping_time}, we define the memorization time 
\begin{equation} 
\tmem(\delta)=\inf\{t>0:\Emem(t)\le\delta\}, 
\label{eq:memory-time}
\end{equation} 
and use $\delta=0.01$ in Figure~\ref{fig:regimes}. This threshold measures when factual classification error (memorization error) becomes small; exact interpolation of the training labels is reached asymptotically. By varying the initialization, sample ratio and kernel angle, we obtain the four behaviors shown in Figure~\ref{fig:regimes}, with agreement between theory and simulation in each case.

In panel (a), facts are memorized before generalization has finished improving. The large initialization delays the emergence of the rule in the predictions. Panel (b) shows a similar delay, followed by substantial deterioration as training fits label noise along the kernel eigenvectors associated with small eigenvalues. These are the slowest modes of gradient flow. Both panels have a finite test optimum, although the late deterioration in (a) is weak. Their delayed generalization resembles grokking \citep{power2022}. Here this separation occurs in the presence of unpredictable labels and requires no feature learning.

In panel (c), the ordering is reversed: optimal stopping occurs while factual memorization is still incomplete. At $\alpha>1$, the sample covariance has a null space whose labels are fitted through the diagonal kernel component. These directions affect factual memorization without contributing directly to the test overlaps, allowing memorization to continue well after the test optimum. This is the classical overfitting scenario: test error increases as training continues to fit the unpredictable labels. In panel (d), useful generalization likewise develops while many facts remain unfitted, but the critical kernel continues to improve until convergence. Its interpolating endpoint therefore attains the best theoretical test error. Together, the four panels show that the theory captures both the different stopping regimes and the separation of learning times within them.

\FloatBarrier \subsection{Kernel-dependent stopping on CIFAR10} \label{app:cifar-numerics} We now examine the same generalization-memorization comparison on CIFAR10 data. Using the official CIFAR10 split \citep{krizhevsky2009}, we construct training sets of $3072$ images with balanced cat and dog classes and approximately $20\%$ automobiles carrying fixed independent random labels. Generalization is measured on the official test cats and dogs, and memorization on the randomly labeled training automobiles. We repeat the experiment with five independently sampled training sets and factual label assignments.

Pixel coordinates are standardized using training statistics, and each image is then normalized to unit Euclidean norm. On these inputs $\widetilde x$, we use the full RBF kernel 
\begin{equation} 
k_\eta(\widetilde x,\widetilde x') =\exp[-\eta\|\widetilde x-\widetilde x'\|_2^2/2] =\exp[-\eta(1-\widetilde x^\top\widetilde x')]. 
\label{eq:cifar-exact-kernel} 
\end{equation} 
We use zero initialization and zero ridge. The RKHS dynamics are evaluated by eigendecomposition on a logarithmic time grid, with the interpolating predictor computed separately at $t=\infty$.

\begin{figure}[!htbp] \centering \includegraphics[width=0.6\linewidth]{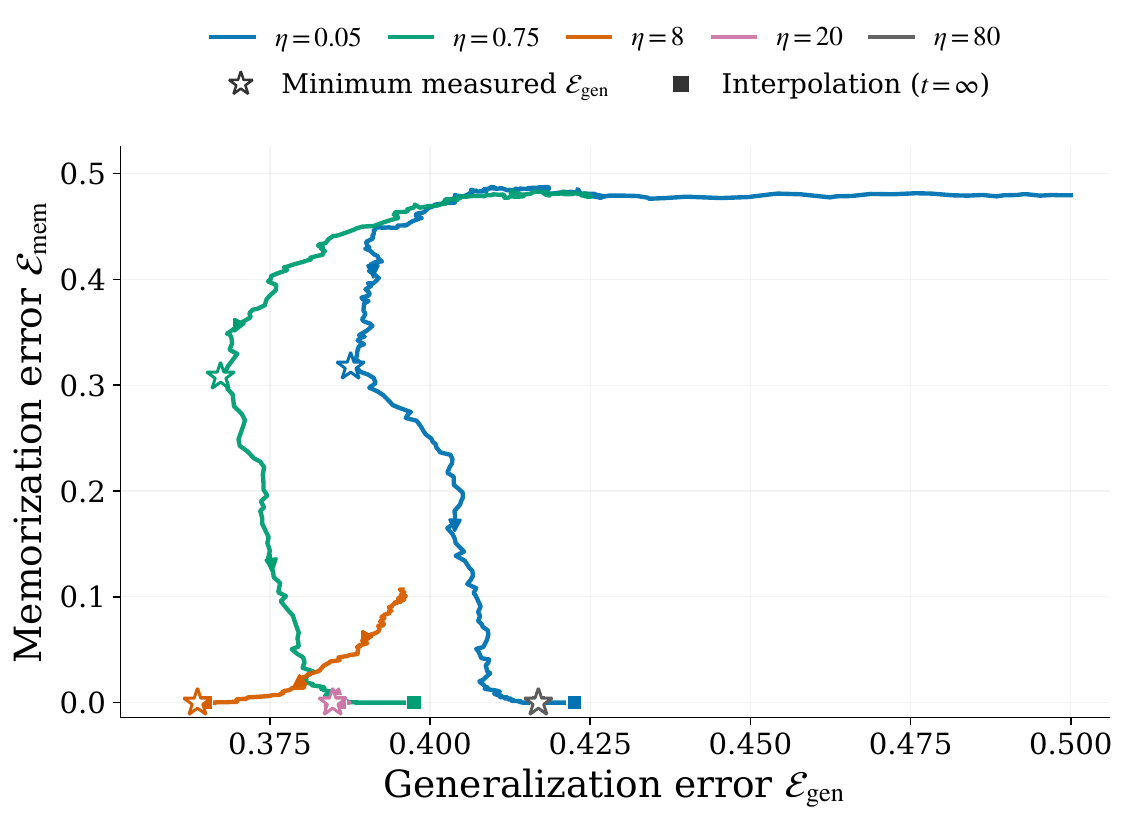} \caption{\textbf{Kernel-dependent generalization and memorization on CIFAR10.} Full RBF-kernel gradient flow with zero initialization and zero ridge. Curves show the unsmoothed mean errors over five repetitions; arrows indicate increasing time. Stars mark the minimum mean test error on the evaluated grid, including the exact convergence endpoint and choosing it when tied. Filled squares mark interpolation. The different bandwidths reproduce the qualitative contrast between trajectories that benefit from early stopping and trajectories with little or no observed deterioration near interpolation.} \label{fig:cifar-kernel} \end{figure}

Figure~\ref{fig:cifar-kernel} shows the mean trajectories for several kernel bandwidths. For some kernels, test error initially decreases and then rises as factual memorization is completed, producing a clear benefit from early stopping. For others, the trajectory reaches interpolation with little or no observed loss of generalization. The resulting endpoint errors nevertheless differ substantially: a kernel with little stopping benefit need not be the one with the best test performance.

This is the same qualitative distinction that appears in the theory. Kernel choice controls both whether late training is harmful and how well the interpolating predictor generalizes. The bandwidths shown were selected for illustration. These image experiments provide a qualitative comparison with the theoretical regimes shown in the main text.

\FloatBarrier

\section{Neural-network experiments on CIFAR10} \label{app:nn}

In this section we describe the experiments behind Figure~\ref{fig:capacity} and examine how the observation varies with the training recipe used. The question is whether fitting the unpredictable labels in the dataset necessarily increases the test error on new cats and dogs with their original labels. We first report the square-loss protocol and unsmoothed network trajectories including the training loss, so as to complement Figure~\ref{fig:capacity}. We then consider augmented cross-entropy training and additional optimizer and residual-network controls. We also randomize labels on cats and dogs themselves, and vary the optimizer, sample size, batch size and random-label fraction. All these experiments compare finite training trajectories.

\subsection{Data, observables and square-loss training} \label{app:square-training} Following the CIFAR10 dataset construction of \citet{farne2026}, the dataset in Figure~\ref{fig:capacity} contains $2800$ cats labelled $+1$, $2800$ dogs labelled $-1$, and $1400$ automobiles assigned independent uniform labels in $\{-1,+1\}$. Thus $n=7000$ and $\varepsilon=0.2$. The official CIFAR-10 test split supplies $1000$ new cats and $1000$ new dogs. Pixels are standardized coordinatewise using training-set statistics. Errors are evaluated on the original, unaugmented images. Batch-normalization layers use the means and variances accumulated during training, which remain fixed during evaluation. A nonnegative score is classified as $+1$.

Alongside clean test error $\Egen$ and factual memorization error $\Emem$ of Equations~\eqref{eq:gen-def} and ~\eqref{eq:mem-def}, we also record the mean squared training loss $L_{\rm MSE}=\frac1n\sum_{i=1}^n(f_\theta(x_i)-y_i)^2$. $\Emem=0$ means that every randomized automobile is classified correctly, while zero square loss requires the scores themselves to equal their labels.

The architecture comparison includes a random-feature convolutional network (RF-CNN), a two-layer convolutional neural network (CNN), deeper CNNs and residual networks. The first two models have the form $f(x)=v^\top\phi_F(x)+c$, where $F$ denotes the convolutional weights. RF-CNN keeps $F$ fixed at its random initialization and trains only the affine readout $(v,c)$; the two-layer CNN trains both. The feature map $\phi_F$ applies average pooling, a $3\times3$ convolution with rectified linear activation (ReLU), spatial averaging, and a $1\times1$ convolution with ReLU.

The four- and six-layer CNNs use blocks comprising a convolution, batch normalization and ReLU, followed by global average pooling, layer normalization and an affine output. ResNet-16 and WRN-28 denote a residual network and a wide residual network, respectively. Their depth labels follow the convention $6B+4$, where $B$ is the number of residual blocks in each of three groups: two for ResNet-16 and four for WRN-28 \citep{zagoruyko2016}. Each block contains two convolutions, with batch normalization and ReLU applied before each convolution (preactivation). A shortcut (a connection bypassing the two convolutions) adds the block input to its output, using a projection (a learned $1\times1$ convolution) when their dimensions differ. Neither network uses dropout, and their widths are adjusted to the prescribed capacity used in each panel. Widths are chosen near $\kappa=P/n=1,10,100$, where $P$ counts the number of all optimized weights for each architecture used. For the RF-CNN, $P$ counts only the readout; the fixed first layer weights are additional.

Trainable networks minimize full MSE with constant-rate Nesterov SGD, momentum $0.9$, batch size $128$. The gradient of each parameter tensor is clipped to Euclidean norm five to limit unusually large updates. There is no augmentation, weight decay or learning-rate decay. The prescribed learning rates range from $0.005$ to $0.02$, depending on family and capacity, and are given in the code. Convolutional weights use Keras He-normal initialization: a zero-centered Gaussian distribution with its tails truncated, rescaled to variance $2/n_{\rm in}$. For a $k_h\times k_w$ convolution with $C_{\rm in}$ input channels, $n_{\rm in}=k_hk_wC_{\rm in}$ counts the scalar values used to compute one output activation. The two-layer CNN's readout weights are independent zero-mean Gaussian variables with variance equal to the inverse feature dimension; the deeper CNNs and residual networks use He-normal initialization for their readouts. Biases start at zero. Each epoch visits every training example in a random order. Training continues for $1500$ epochs, corresponding to $82500$ minibatch updates.

We use a fixed dataset and vary the initialization and minibatch order across seeds. Each curve averages over $28$-$30$ completed runs. The dataset is fixed in all these runs.

The random feature CNN readouts start from zero and follow full-batch GD on MSE. Features are standardized using training statistics, scaled by the inverse square root of their dimension and supplemented with an intercept. The step size is $0.9n/(2\sigma_{\max}^2)$, where $\sigma_{\max}$ is the largest singular value of this design matrix. The finite GD iterates are evaluated spectrally through $10^{12}$ steps, using a feature SVD or a Gram eigendecomposition according to feature dimension.

\begin{figure}[!htbp] \centering \includegraphics[width=1\linewidth]{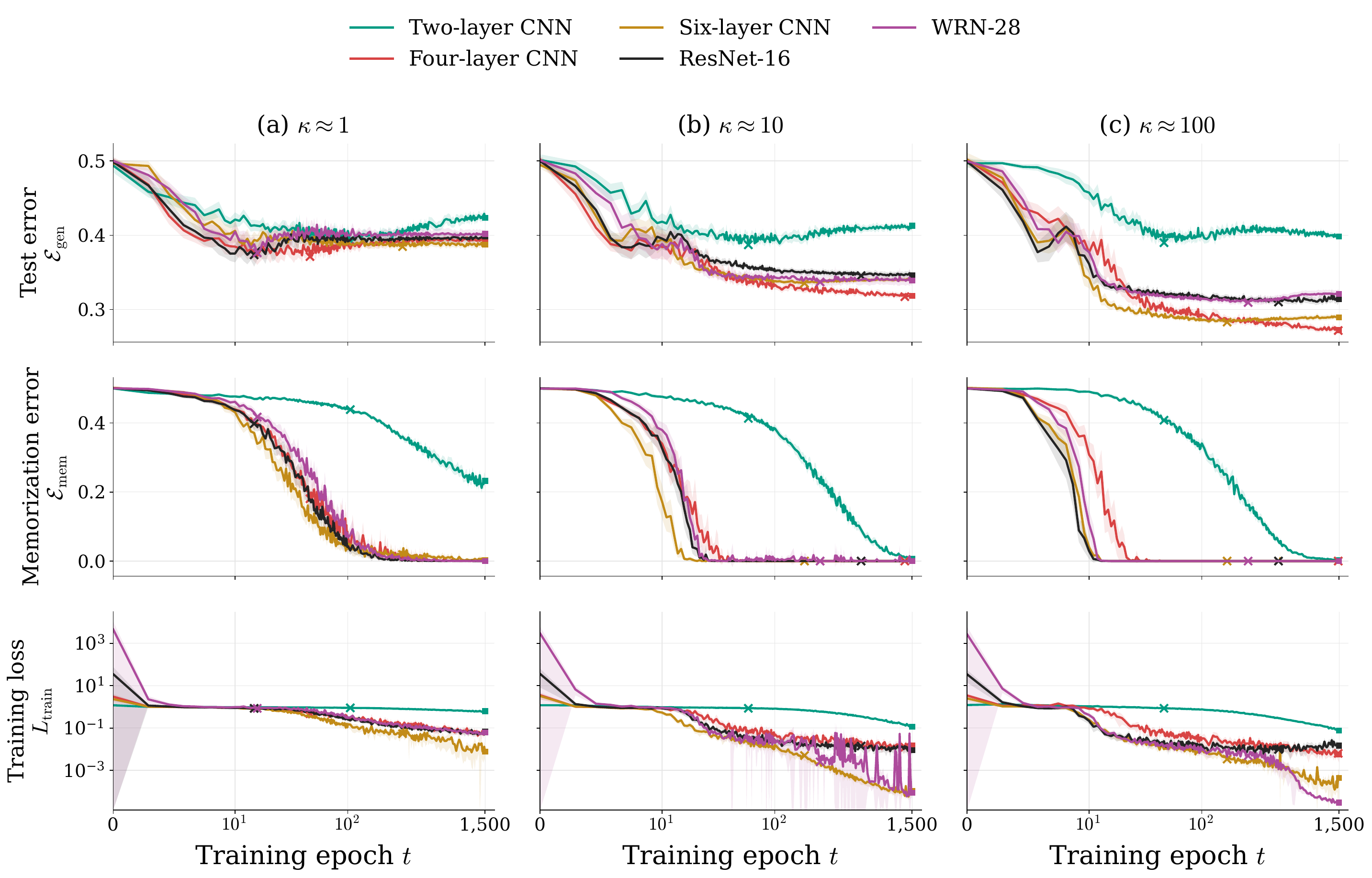} \caption{\textbf{Unsmoothed dynamics behind Figure~\ref{fig:capacity}.} Test error, memorization error and full training MSE (top to bottom), averaged over $28$--$30$ completed seeds per curve through $1500$ epochs at $\kappa\approx1,10,100$. Shading gives pointwise $95\%$ confidence intervals for the mean across seeds. Crosses mark each raw mean test curve's minimum and the corresponding values of the other metrics; squares mark the last observation. No temporal averaging is applied. The models are two-, four- and six-layer CNNs, a residual network with nominal depth $16$ (ResNet-16), and a wide residual network with nominal depth $28$ (WRN-28).} \label{fig:capacity-time-cnn} \end{figure}

For readability, Figure~\ref{fig:capacity} averages each error coordinate over at most seven neighboring saved observations within each seed, then averages across seeds. The window shrinks symmetrically near the boundaries, so initialization and the final iterate are preserved. Figure~\ref{fig:capacity-time-cnn} therefore reports the corresponding test error, memorization error and training MSE without temporal smoothing. Lines show means across seeds, and shading gives pointwise $95\%$ confidence intervals for these means. Crosses use the minimum raw mean test error, and squares mark the last observation. Time is displayed as $\log_{10}(1+t)$ with epoch labels; the loss axis is logarithmic.

The larger deep networks retain good test performance as factual error approaches zero, while the two-layer CNN remains less accurate. The loss panels also show why factual classification and exact interpolation should not be conflated: the observed training MSE remains positive, and keeps decreasing even if moderately after $1500$ epochs, except with some numerical spikes. The empirical observation is that completing factual classification, i.e. achieving zero memorization error, need not produce a substantial late loss of generalization. Small numerical fluctuations and a finite gap between the best recorded and final test errors remain compatible with that observation.

 \FloatBarrier

\subsection{Augmented cross-entropy training} \label{sec:experiments}

We next ask whether the same observation persists under a training recipe commonly used to obtain good test performance on CIFAR10: cross-entropy, random crops and horizontal flips, and a scheduled learning rate \citep{zagoruyko2016,loshchilov2017,he2019tricks}. We use a separate fixed dataset containing $2458$ cats, $2458$ dogs and $1228$ randomly labelled automobiles ($n=6144$), with five training seeds and the same $2000$ official-test cats and dogs. Inputs are standardized using this dataset's training statistics.

4-layer CNN, 6-layer CNN, ResNet-16, WRN-28 and MobileNetV2 each have approximately $2.7$ million optimized parameters. Here, the model labelled ResNet-16 is the wide residual architecture WRN-16-4: the width multiplier four gives $64$, $128$ and $256$ channels in its three residual groups. MobileNetV2 is a computationally efficient convolutional architecture that combines convolutions acting separately on each channel with $1\times1$ convolutions that mix channels \citep{Sandler_2018_CVPR}. We adapt its first convolution to preserve the spatial resolution of the small CIFAR10 images, using stride one, and replace its classifier with a scalar output. All models are trained from scratch.

Each epoch presents one randomly transformed view of every training image: four-pixel zero padding, a random $32\times32$ crop and a horizontal flip with probability $1/2$. Images are standardized before augmentation, and labels remain fixed. Transformations are paired across architectures by seed, image and epoch. We minimize mean binary cross-entropy, 
\begin{equation} 
L_{\rm BCE}=\frac1n\sum_{i=1}^n \log\!\left(1+\exp[-y_i f_\theta(x_i)]\right), 
\label{eq:bce-loss} 
\end{equation} 
using Nesterov SGD with momentum $0.9$, batch size $128$ and gradient clipping at norm five for each parameter tensor, as in the square-loss experiments above. The learning rate increases linearly to $0.05$ over five epochs, follows a cosine decay to $0.0005$ at epoch $400$, and remains at this positive value thereafter. The schedule is specified in advance, independently of test performance or training classification error. There is no weight decay, dropout or label smoothing. All reported errors and losses are evaluated on the original, unaugmented images.

The reported trajectories extend to $2000$ epochs, except for WRN-28, which is shown through $1000$ epochs. Across all five architectures (the four and six-layer CNNs, ResNet-16, WRN-28 and MobileNetV2), the test error settles onto an approximately constant plateau with zero memorization error, while training cross-entropy continues to decrease. Although the attained test errors differ across architectures, continued training preserves performance close to the best observed level, with only small fluctuations and little visible advantage from early stopping. Thus, even under this recipe, completing factual memorization does not lead to a substantial deterioration in generalization. The continued decrease in cross-entropy reflects increasing classification margins after the random labels have already been classified correctly.

\begin{figure}[!htbp] \centering \includegraphics[width=.60\linewidth]{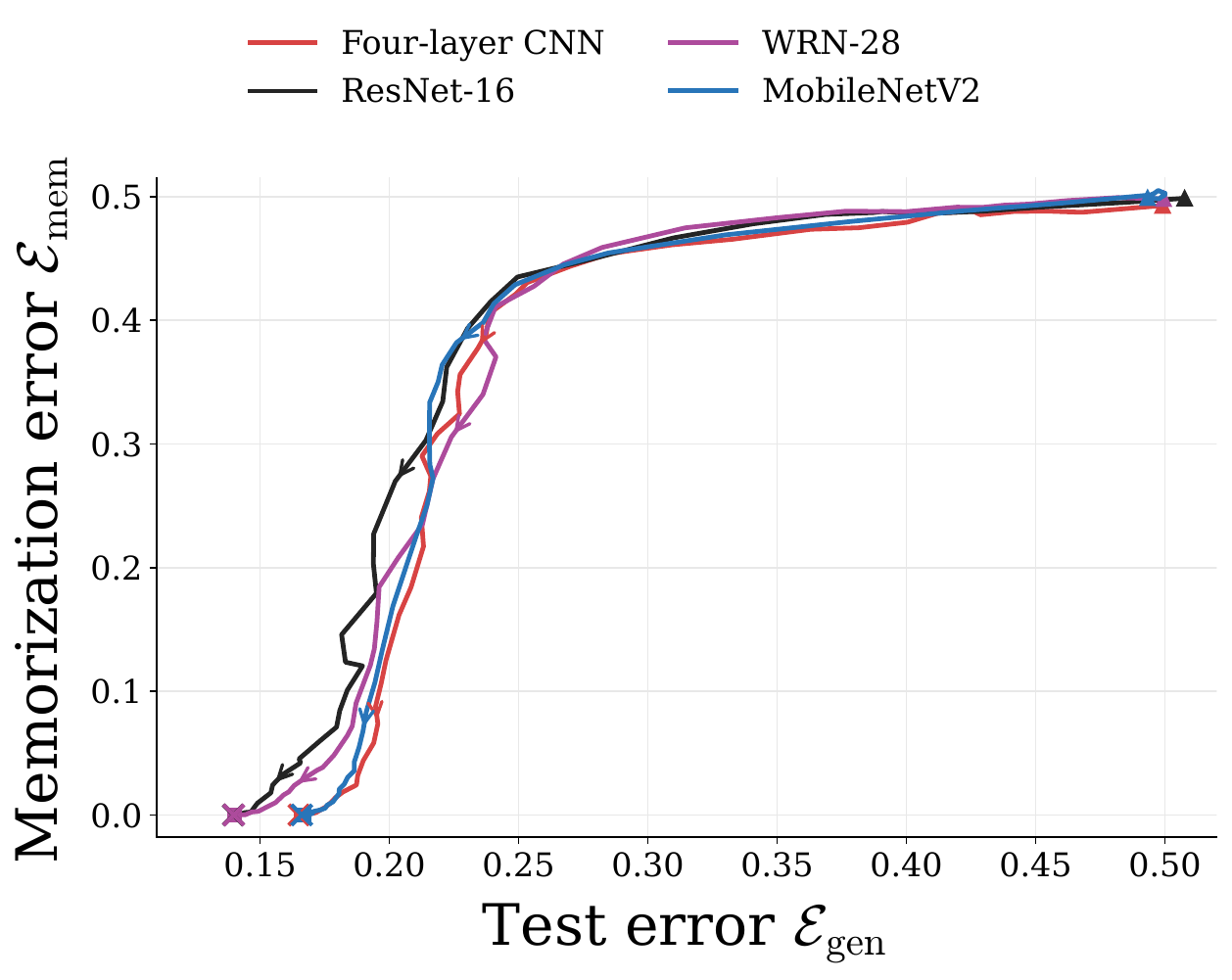} \caption{\textbf{Generalization while fitting random labels with augmented cross-entropy.} Clean cat-dog test error versus classification error on randomly labelled automobiles, for the $6144$-image dataset and approximately $2.7$ million optimized parameters per network. Coordinates are averaged over at most five neighboring saved observations within each seed, then over five seeds. Triangles mark initialization, arrows indicate increasing time, crosses mark the minimum displayed mean test error and squares mark the reporting endpoints. Trajectories extend to $2000$ epochs, except WRN-28, which is shown through $1000$ epochs.} \label{fig:empirical} \end{figure}

\begin{figure}[!htbp] \centering \includegraphics[width=1.\linewidth]{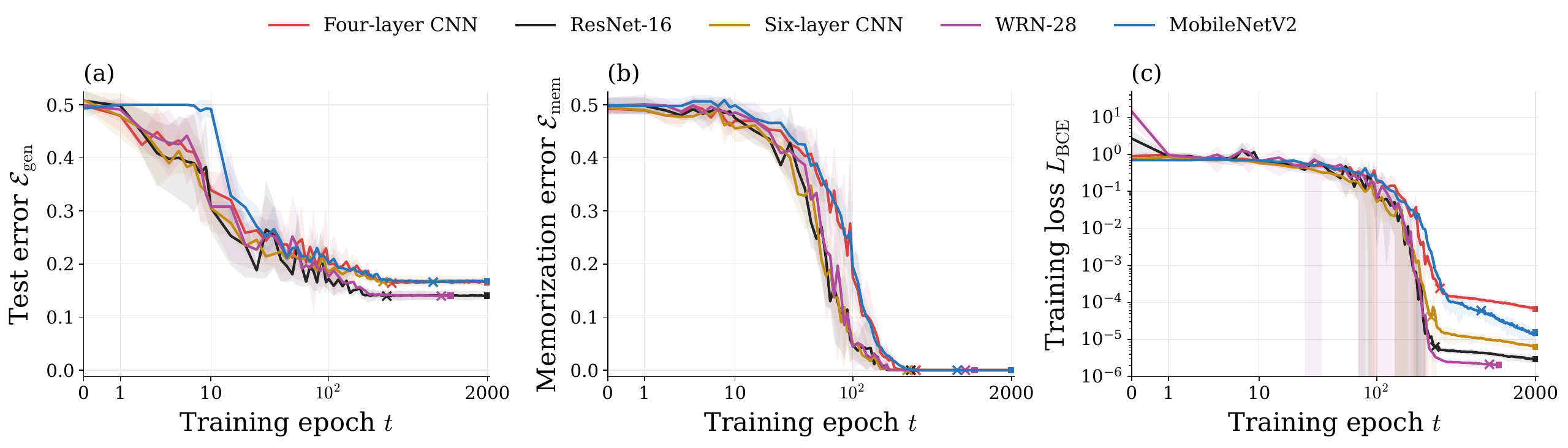} \caption{\textbf{Unsmoothed trajectories under augmented cross-entropy training.} Clean test error, automobile memorization error and mean binary cross-entropy on all original training images. Lines show five-seed means; shading gives pointwise $95\%$ confidence intervals for these means. Crosses mark the minimum raw mean test error and the corresponding values of the other metrics; squares mark the reporting endpoints. The comparison includes CNN4, ResNet-16, CNN6, WRN-28 and MobileNetV2. Trajectories extend to $2000$ epochs, except WRN-28, which is shown through $1000$ epochs. Time is displayed as $\log_{10}(1+t)$ with epoch labels; the cross-entropy axis is logarithmic.} \label{fig:bce-time} \end{figure}

Figure~\ref{fig:empirical} shows the parametric paths, averaging at most five neighboring observations within each seed before taking the seed mean, with endpoints unchanged. Figure~\ref{fig:bce-time} gives the unsmoothed trajectories for all five architectures, including the training cross-entropy. These experiments show that the observation is also present under augmented cross-entropy training. The training horizons are fixed.

 \FloatBarrier

\subsection{Optimizer and residual-network controls} \label{app:controls}

\begin{figure}[!htbp] \centering \includegraphics[width=\linewidth]{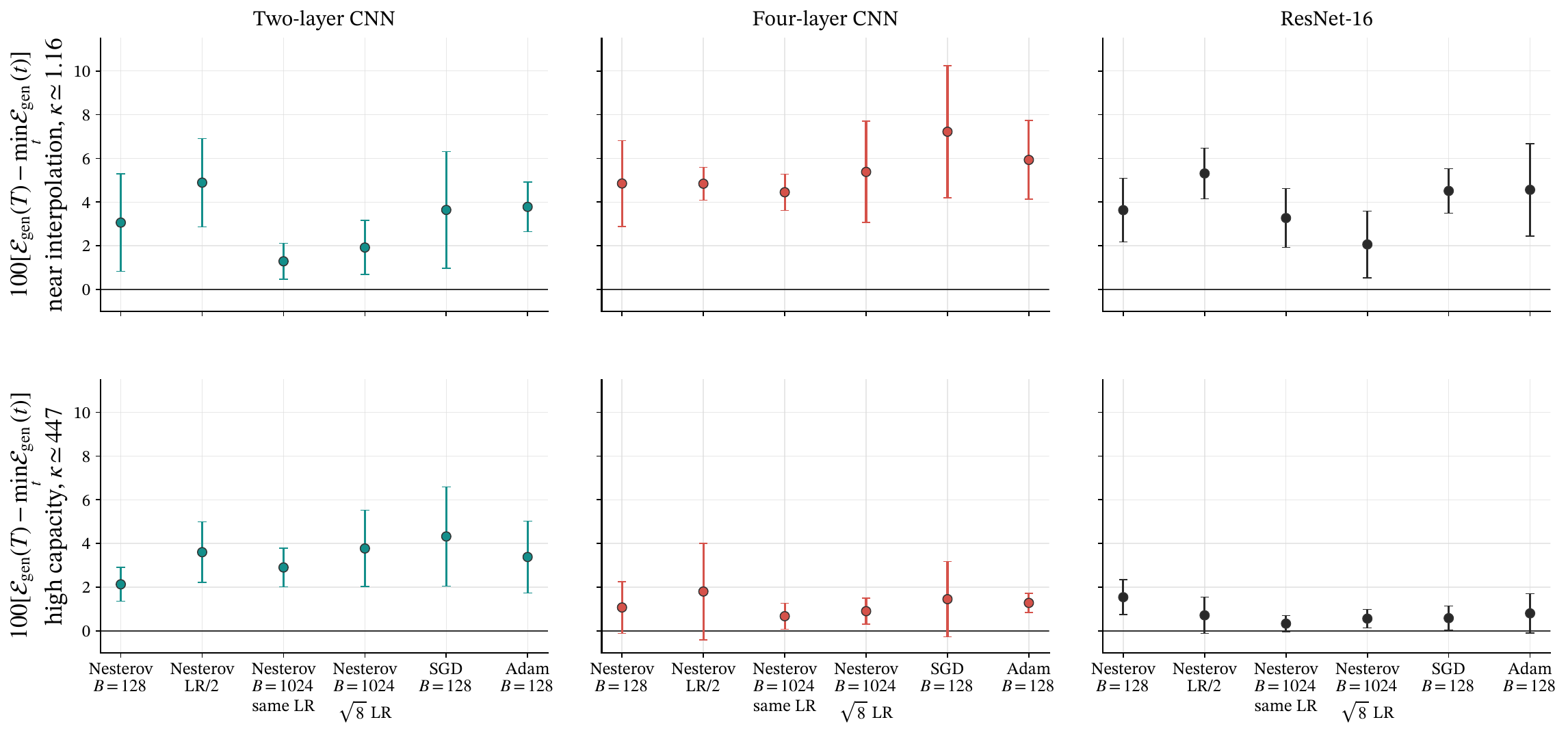} \caption{\textbf{Sensitivity to the optimization protocol.} Difference between the final test error and the lowest test error recorded during each run, averaged over five seeds and expressed in percentage points. Results compare two-layer CNNs, four-layer CNNs and ResNet-16 at intermediate capacity (top) and high capacity (bottom). Error bars give $95\%$ confidence intervals for the mean. The larger deep models retain a smaller gap across the displayed optimizer, learning-rate and batch-size choices.} \label{fig:optimizer} \end{figure}

In this subsection we examine whether the architectural comparison is sensitive to the optimizer, learning rate or batch size. Figure~\ref{fig:optimizer} uses the fixed $6144$-image dataset, square loss without augmentation and five seeds. We compare two-layer CNNs, four-layer CNNs and ResNet-16 at $\kappa= P/n\simeq1.16$ and $447$.

Starting from the baseline Nesterov protocol, we halve the learning rate or increase the batch size from $128$ to $1024$. For the larger batch, we either retain the baseline rate or multiply it by $\sqrt8$. We also consider SGD without momentum at five times the baseline rate and Adam at learning rate $10^{-3}$, both with batch size $128$. Nesterov momentum remains $0.9$, all learning rates are constant, and gradients are clipped at norm five for each parameter tensor. The CNNs are trained for $800$ epochs and ResNet-16 for $400$.

For each run, we measure the difference between the final test error and its lowest recorded value along that trajectory, then average this difference across seeds. The plot shows that the larger ($\kappa=P/n\approx 447$) deep models show a smaller gap across the displayed protocols. Although the measured gaps vary, the qualitative architectural comparison remains similar under these changes to the optimization procedure.

\begin{figure}[!htbp] \centering \includegraphics[width=0.90\linewidth]{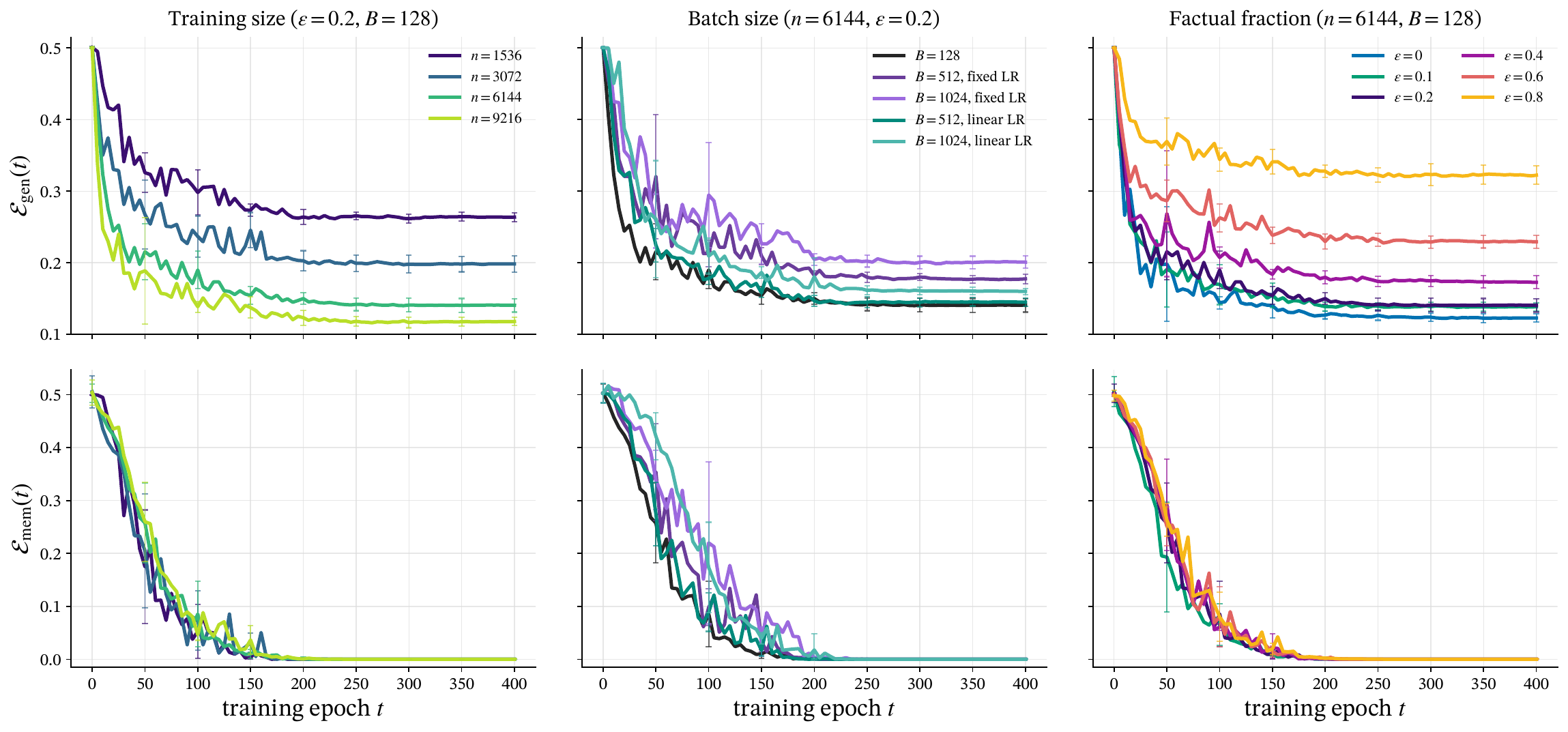} \caption{\textbf{Extended ResNet controls with augmented cross-entropy.} Columns vary training-set size, batch size and random-label fraction under the $400$-epoch recipe of Section~\ref{app:controls}. Means are unsmoothed; bars show one sample standard deviation over ten seeds at $\varepsilon=0.1,0.2$ and five otherwise. ``Fixed LR'' and ``linear LR'' refer to peak rates. Memorization error is omitted at $\varepsilon=0$.} \label{fig:followup-bce-controls} \end{figure}

We then return to augmented cross-entropy training and consider the same ResNet-16 architecture (WRN-16-4) used in Section~\ref{sec:experiments}. Figure~\ref{fig:followup-bce-controls} varies the training-set size, batch size and fraction of random labels over five seeds, with ten seeds at $\varepsilon=0.1,0.2$. The datasets are constructed separately from the preceding experiments using common, fixed pools of cats, dogs and automobiles. In the training-size comparison, each smaller dataset is contained in the larger ones, and shared automobiles retain the same random labels.

Starting from $(n,B,\varepsilon)=(6144,128,0.2)$, where $B$ is the batch size, we vary $n$ from $1536$ to $9216$, $B$ from $128$ to $1024$, and $\varepsilon$ from $0$ to $0.8$, changing one quantity at a time. For larger batches, the peak learning rate is either held at $0.05$ or increased in proportion to the batch size. Training uses the augmentation, loss, momentum and gradient clipping described in Section~\ref{sec:experiments}, with five epochs of warm-up followed by cosine decay to $1\%$ of the peak rate at epoch $400$. These runs stop at epoch $400$, without the subsequent constant-rate tail. Errors on the original images are evaluated every five epochs.

\begin{figure}[!htbp] \centering \includegraphics[width=0.90\linewidth]{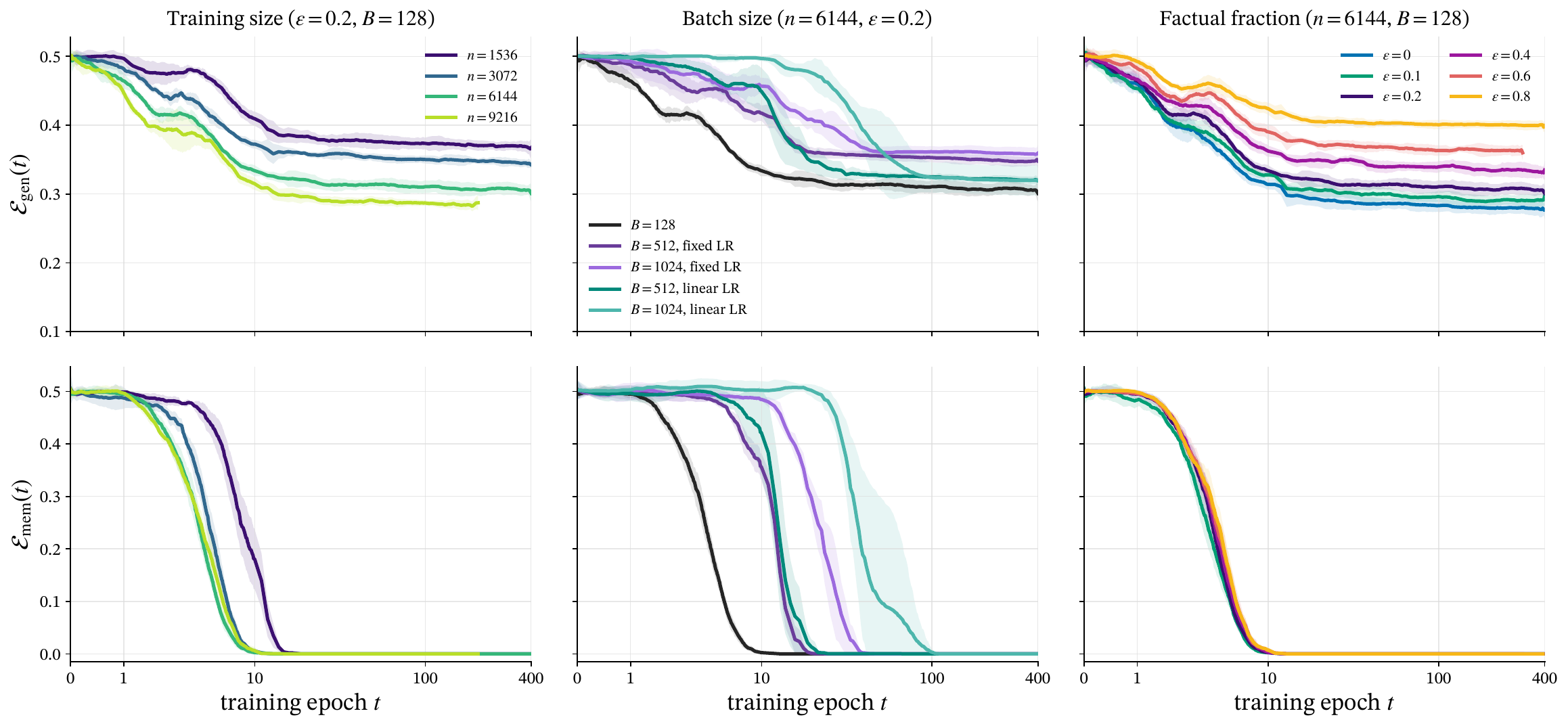} \caption{\textbf{ResNet controls with constant-rate square loss.} Conditions follow Figure~\ref{fig:followup-bce-controls}, with five seeds. Per-seed arithmetic averaging uses a symmetric window of half-width at most $0.08$ in $\log_{10}(1+t)$, shrinking at the endpoints. Lines and shading give means and one sample standard deviation. Horizons are $200$ epochs for $n=9216$, $300$ for $\varepsilon=0.6$, and $400$ otherwise; time is labelled in epochs.} \label{fig:followup-square-controls} \end{figure}

Figure~\ref{fig:followup-square-controls} repeats these conditions with square loss, no augmentation and constant learning rates: $0.01$, or $0.04$ and $0.08$ under linear scaling at batch sizes $512$ and $1024$. The remaining settings are unchanged.

Across both recipes, all displayed conditions with random labels reach zero memorization error, while test performance approaches a plateau close to its lowest recorded value and remains stable through the end of training. The level of this plateau nevertheless varies substantially. Increasing the training-set size also changes $P/n$, while increasing the random-label fraction at fixed $n$ leaves fewer rule examples. Thus, the quality of generalization changes across these conditions, but completing factual memorization does not produce a substantial late deterioration. The observation therefore persists beyond a single choice of sample size, batch size or random-label fraction. Memorization error is undefined at $\varepsilon=0$ and is omitted from the plot.

The captions specify the reporting horizons, including the two square-loss conditions shown before their planned $400$ epochs. These comparisons concern the complete training recipes and do not equate zero classification error with zero loss.

\FloatBarrier \Needspace{6\baselineskip} \subsection[Random labels on cats and dogs]{Random labels on cats and dogs} \label{app:followup}

We now ask whether the preceding observations depend on using a separate image class for the random labels. We replace the automobiles by a disjoint subset of cats and dogs with independent uniform labels in $\{-1,+1\}$. The square-loss dataset contains $5600$ clean and $1400$ randomized examples; the cross-entropy dataset contains $4916$ clean and $1228$ randomized examples, balanced between the two classes in each subset. We retain the clean training and test images from the automobile experiments and recompute training normalization statistics. Data and labels are fixed across seeds. The observables follow Eqs.~\eqref{eq:gen-def} and~\eqref{eq:mem-def}; the randomized subset includes labels that happen to agree with the original class.

\begin{figure}[!htbp] \centering \includegraphics[width=\linewidth]{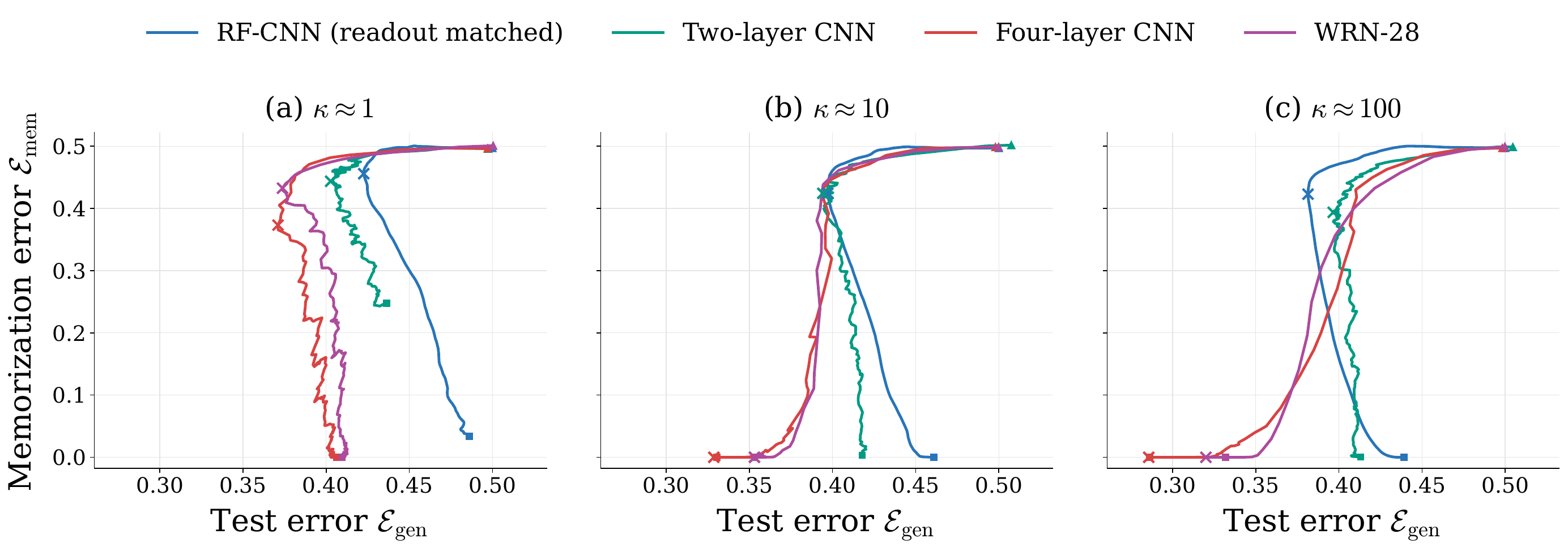} \caption{\textbf{Memorization paths with random labels on cats and dogs.} Square-loss training at $\kappa=P/n\approx1,10,100$, averaged over ten seeds. Both coordinates use the endpoint-preserving average over at most seven saved times described in Section~\ref{app:square-training}. Triangles mark initialization, crosses the minimum displayed mean test error and squares the endpoints. Trainable networks run for $1500$ epochs; RF readout GD runs through $10^{12}$ full-batch steps.} \label{fig:followup-square} \end{figure}

\begin{figure}[!htbp] \centering \includegraphics[width=\linewidth]{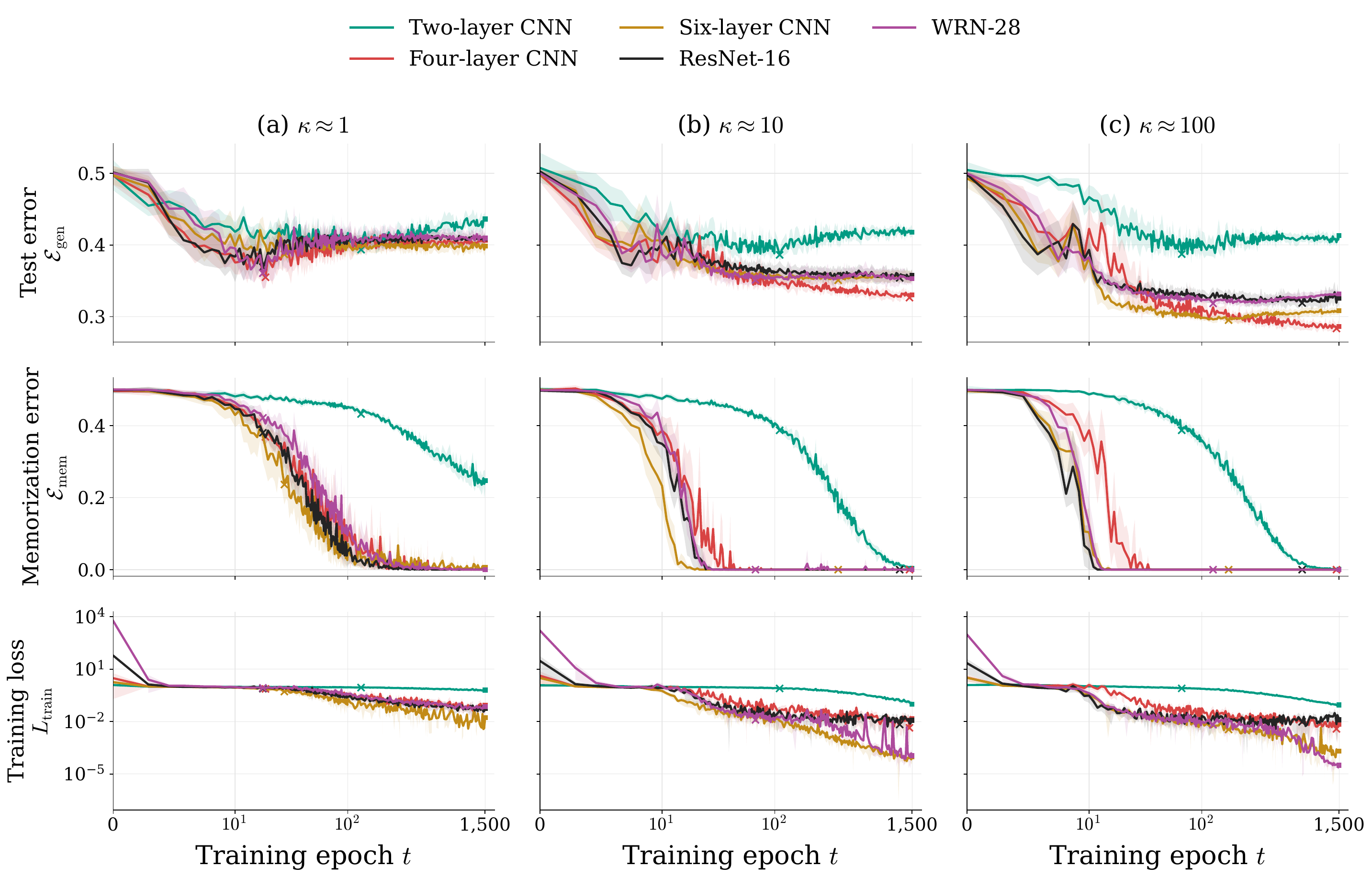} \caption{\textbf{Unsmoothed square-loss trajectories with random labels on cats and dogs.} Test error, memorization error and training square loss for the five trainable architectures, through $1500$ epochs. Lines show ten-seed means; shading gives pointwise $95\%$ confidence intervals. Crosses mark the minimum raw mean test error and corresponding other metrics; squares mark endpoints. Time uses $\log_{10}(1+t)$ with epoch labels, and loss is shown logarithmically.} \label{fig:followup-square-time} \end{figure}

The square-loss protocol follows Section~\ref{app:square-training}. Figures~\ref{fig:followup-square} and~\ref{fig:followup-square-time} show that, at the two larger capacities, the deeper trainable networks reach their lowest mean test error after memorization error reaches zero. Subsequent deterioration is smaller than for frozen features and the two-layer CNN, although some architectures retain a visible gap.

\begin{figure}[!htbp] \centering \includegraphics[width=0.60\linewidth]{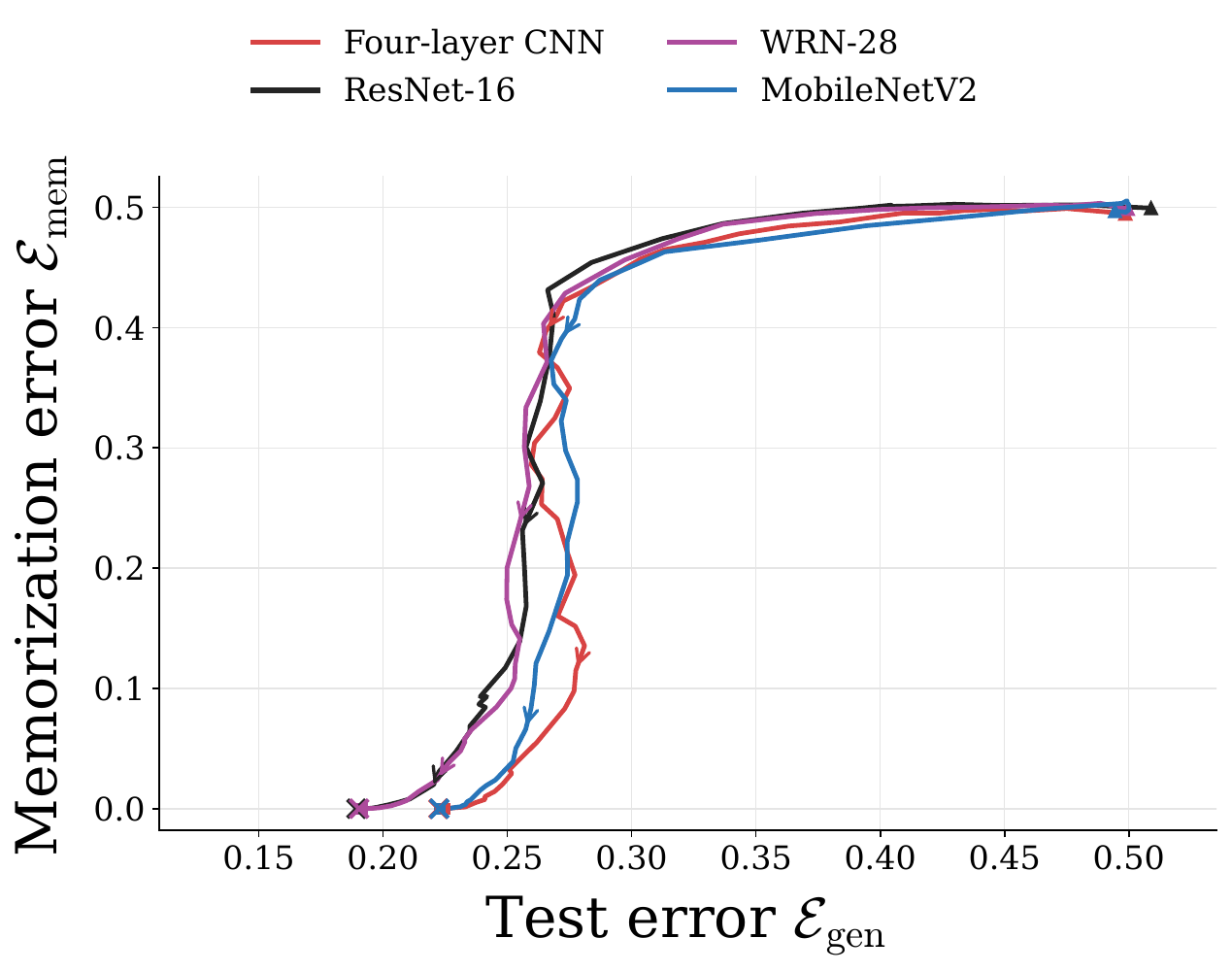} \caption{\textbf{Memorization paths with augmented cross-entropy and randomized cats and dogs.} Five-seed means after endpoint-preserving averaging over at most five saved times; markers follow Figure~\ref{fig:followup-square}. Reporting horizons are $2000$ epochs for CNN4, $1100$ for ResNet-16, $900$ for WRN-28 and $1700$ for MobileNetV2. Each network has approximately $2.7$ million optimized parameters.} \label{fig:followup-bce} \end{figure}

\begin{figure}[!htbp] \centering \includegraphics[width=\linewidth]{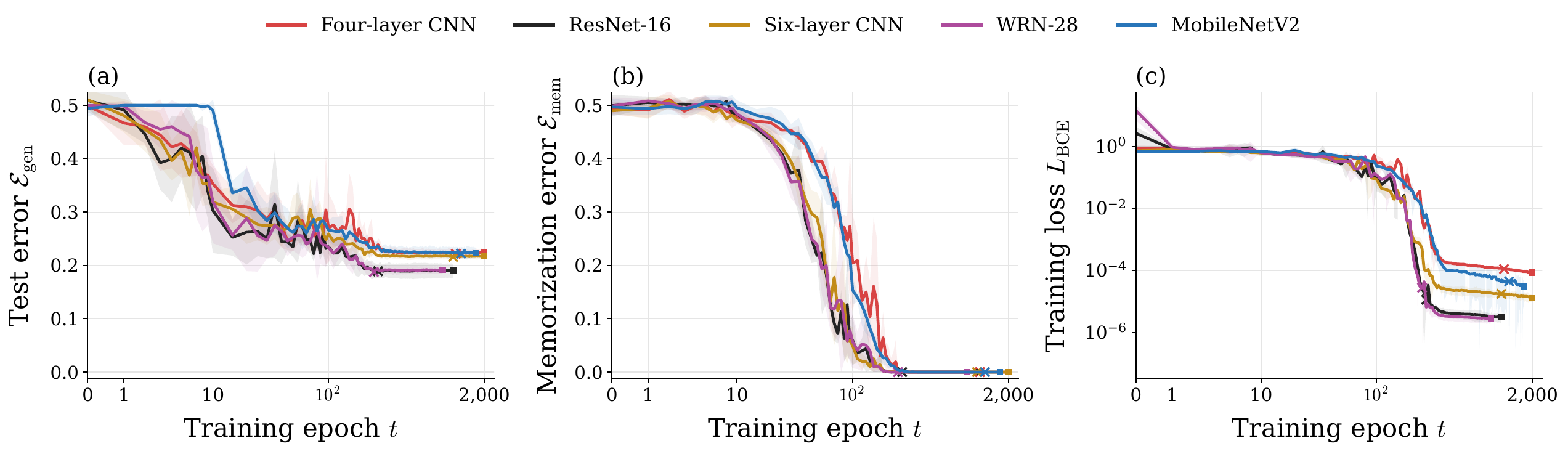} \caption{\textbf{Unsmoothed cross-entropy trajectories with randomized cats and dogs.} Five-seed means of test error, memorization error and training cross-entropy on original images, with pointwise $95\%$ confidence intervals. Horizons follow Figure~\ref{fig:followup-bce}; CNN6 is additionally shown through $2000$ epochs. Markers and axes follow Figure~\ref{fig:followup-square-time}. No temporal averaging is applied.} \label{fig:followup-bce-time} \end{figure}

With the augmented cross-entropy recipe of Section~\ref{sec:experiments} and Eq.~\eqref{eq:bce-loss}, all five architectures reach zero memorization error while maintaining test performance close to their lowest observed mean error (Figures~\ref{fig:followup-bce} and~\ref{fig:followup-bce-time}). Training cross-entropy continues to decrease. Thus, the observation also holds when clean and randomized examples belong to the same image classes.

For further details of the numerical implementations, we refer to the repository at \url{https://github.com/SPOC-group/Early-stopping}.

 \FloatBarrier

\end{document}